\documentclass[oneside,final, letterpaper]{ucr}
\DeclareUnicodeCharacter{2217}{ }
\begin{document}


\title{Time Series Data Mining Algorithms Towards Scalable \\and Real-Time Behavior Monitoring}
\author{Alireza Abdoli}
\degreemonth{September}
\degreeyear{2021}
\degree{Doctor of Philosophy}
\chair{Dr. Eamonn Keogh}
\othermembers{Dr. Vagelis Papalexakis\\
Dr. Tamar Shinar\\
Dr. Jiasi Chen}
\numberofmembers{4}
\field{Computer Science}
\campus{Riverside}

\maketitle
\copyrightpage{}
\approvalpage{}

\degreesemester{Summer}

\begin{frontmatter}

\begin{acknowledgements}
I am grateful to my wonderful advisor, Distinguished Professor Eamonn Keogh without his help, I would not have been here. Also, I am grateful to the opportunity to work with Professor Alec Gerry and Professor Amy Murillo from Entomology department at the University of California, Riverside.

To my parents and brother who have supported me in every possible way. I am blessed to have you by my side.

Much of the work in this dissertation has already been published in conferences sponsored by the Institute of Electrical and Electronics Engineers (IEEE) and the Association for Computing Machinery (ACM) from the following published works:

\begin{itemize}
    \item IEEE International Conference on Big Data (Chapter 2)
    \item IEEE International Conference on Machine Learning and Applications (Chapter 3)
    \item SIGKDD Conference on Knowledge Discovery and Data Mining (Chapter 4)
\end{itemize}

\end{acknowledgements}

\begin{dedication}
\null\vfil
{\large
\begin{center}
To my savior who has been supporting me through all the ups and downs of my life. Please don't ever let go of my hand.
\end{center}}
\vfil\null
\end{dedication}

\begin{abstract}
In recent years, there have been unprecedented technological advances in sensor technology, and sensors have become more affordable than ever. Thus, sensor-driven data collection is increasingly becoming an attractive and practical option for researchers around the globe. Such data is typically extracted in the form of time series data, which can be investigated with data mining techniques to summarize the behaviors of a range of subjects including humans and animals. While enabling cheap and mass collection of data, continuous sensor data recording results in datasets which are big in size and volume, which are challenging to process and analyze with traditional techniques and software in a timely manner. Such collected sensor data is typically extracted in the form of time series data.

Time series data is widely used in many domains and is of significant interest in data mining. In recent years researchers have proposed various algorithms for the efficient processing of time series datasets. The problem of time series classification has been around for decades. There are two main approaches for time series classification in the literature, namely, shape-based classification and feature-based classification. Shape-based classification determines the best class according to a distance measure (e.g., Euclidean distance). Feature-based classification, on the other hand, measures properties of the time series and finds the best class according to the set of features defined for the time series.

In this dissertation, we demonstrate that neither of the two techniques will dominate for some problems, but that some “combination” of both might be the best. In other words, on a single problem, it might be possible that one of the techniques is better for one subset of the behaviors, and the other technique is better for another subset of behaviors. We introduce a hybrid algorithm to classify behaviors, using both shape and feature measures, in weakly labeled time series data collected from sensors to quantify specific behaviors performed by the subject. We demonstrate that our algorithm can robustly classify real, noisy, and complex datasets, based on a combination of shape and features, and tested our proposed algorithm on real-world datasets.
\end{abstract}

\tableofcontents
\listoffigures
\listoftables
\end{frontmatter}


\chapter{Introduction}

In recent years, there have been unprecedented technological advances in sensor technology, and sensors have become more affordable than ever. Sensors can be used to greatly increase the number of individuals that can be tracked, while also expanding the tracking period, in some cases to 24/7 monitoring. Consequently, sensor-driven data collection is increasingly becoming an attractive and practical option for researchers around the globe. Such collected sensor data is typically extracted in the form of time series data, which can be investigated with data mining techniques to summarize the behaviors of a range of subjects including humans and animals.
 
While enabling cheap and mass collection of data, continuous sensor data recording results in datasets which are big in size and volume; consequently, such datasets are challenging to process and analyze with traditional techniques and software in a timely manner. Such collected sensor data is typically extracted in the form of time series data, which can be examined with data mining techniques to quantify behaviors of the subject under study.

\section{Advances in Time Series Data Mining}

Time series data is widely used in many domains and is of significant interest in data mining. In recent years researchers have proposed various algorithms for the efficient processing of time series datasets. The problem of time series classification has been around for decades. There are two main approaches for time series classification in the literature, namely, shape-based classification and feature-based classification. Shape-based classification determines the best class according to a distance measure (e.g. Euclidean distance). Feature-based classification, on the other hand, finds the best class according to the set of features defined for the time series. These features measure properties of the time series (e.g. autocorrelation, complexity, etc.).

\section{Contribution}

In this dissertation, we demonstrate that neither of the two techniques will dominate for some problems (i.e. data is unlikely to yield to a single modality of classification), but that some “combination” of both might be the best. In other words, on a single problem, it might be possible that one of the techniques is better for one subset of the behavior, and the other technique is better for another subset of behaviors. We introduce a hybrid algorithm to classify behaviors, using both shape and feature measures, in weakly labeled time series data collected from sensors to quantify specific behaviors performed by the subject. We demonstrate, with an extensive empirical study, that our algorithm can robustly classify real, noisy, and complex datasets, based on a combination of shape and features, and tested our proposed algorithm on real-world datasets.

\section{List of Publications}

Below, comes the list of publications in Institute of Electrical and Electronics Engineering (IEEE) and the Association for Computing Machinery (ACM), that will be discussed throughout the chapters of this dissertation. I am grateful to the IEEE and ACM for the opportunity to include the following works into my dissertation.

\begin{figure}[h]
\centering
\includegraphics[width=\textwidth]{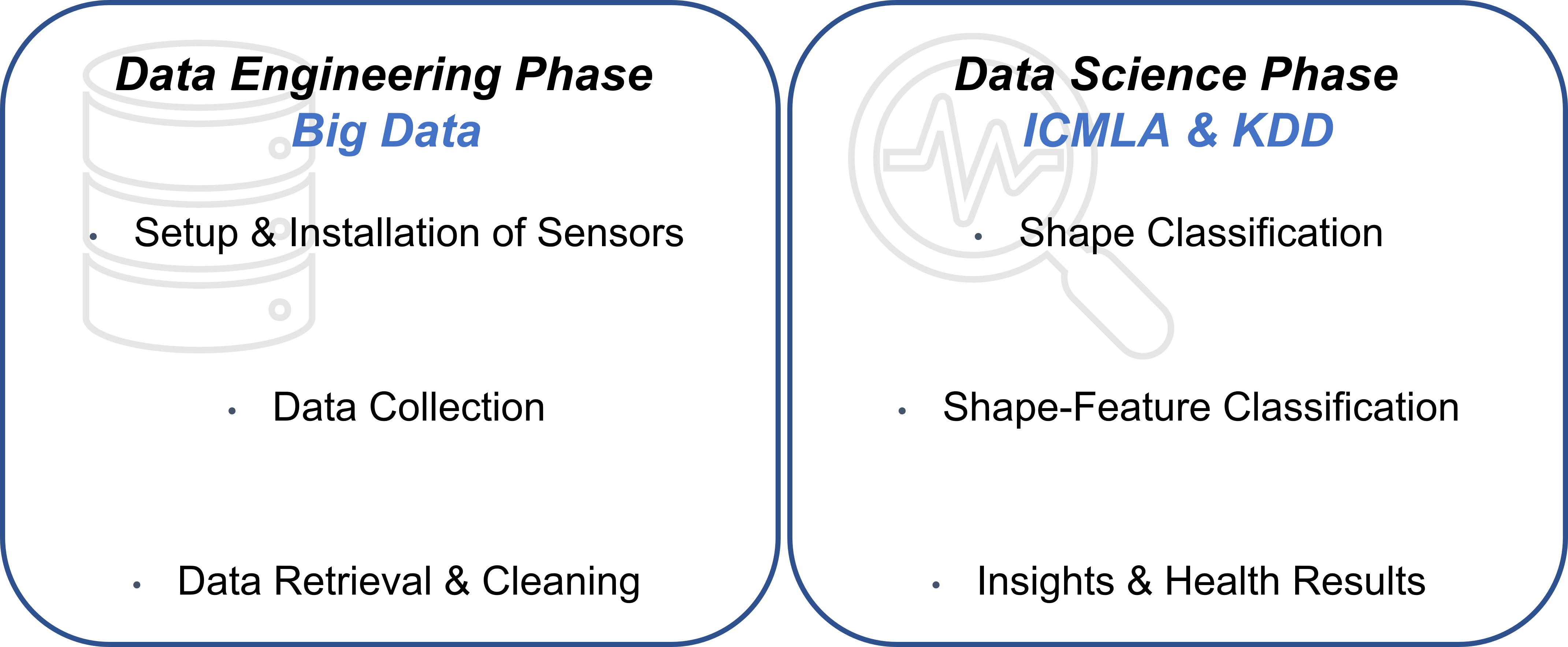}
\caption{Overview of topics to be discussed in this dissertation.}~\label{fig:Intro}
\end{figure}

\begin{enumerate}
    \item {\textbf{Abdoli, A.}, Murillo, A. C., Gerry, A. C., and Keogh, E. J. (2019, December). \textbf{Time Series Classification: Lessons Learned in the (Literal) Field while Studying Chicken Behavior.} In 2019 IEEE International Conference on Big Data (Big Data) (pp. 5962-5964). IEEE.}
    
    \item {\textbf{Abdoli, A.}, Murillo, A. C., Yeh, C. C. M., Gerry, A. C., and Keogh, E. J. (2018, December). \textbf{Time series classification to improve poultry welfare.} In 2018 17TH IEEE International Conference on Machine Learning and Applications (ICMLA) (pp. 635-642). IEEE.}
    
    \item {\textbf{Abdoli, A.}, Alaee, S., Imani, S., Murillo, A., Gerry, A., Hickle, L., and Keogh, E. (2020, August). \textbf{Fitbit for chickens? time series data mining can increase the productivity of poultry farms.} In Proceedings of the 26th ACM SIGKDD International Conference on Knowledge Discovery \& Data Mining (pp. 3328-3336).}
\end{enumerate}
\newpage
The course of this dissertation is shown in Figure \ref{fig:Intro}. Initially, the data engineering part is discussed which involves setting up and installing sensors for data collection, retrieval, and pre-processing (i.e., cleaning and preparation) towards downstream classification algorithms. Simply put, data engineering refers to various algorithms working on collected data to cleanse and divide data into 24-hour chunks and save the data back to the disk.

During the downstream processing, firstly the shape classification algorithm is discussed which works based on a distance measure (e.g., euclidean distance). Next, the superior shape-feature classification approach will be discussed which applies shape classification for some behaviors, while utilizing feature classification, using statistical and mathematical features, for classifying other behaviors. Then, the applicability of the shape-feature classification is demonstrated on chicken data to assess and monitor the overall health status of chickens based on changes in the frequency and timing of the chicken behaviors. Finally, the dissertation is concluded with the future works and research directions.

\chapter{Data Collection \& Preparation: Time Series Classification: Lessons Learned in the (Literal) Field while Studying Chicken Behavior}

\newpage

\section{Introduction}

There was a time, not that long ago, when wearable devices were only the province of humans. How far we have come since then. There are high-tech wearables available, regardless of whether you’re a tiny insect \cite{batista2011sigkdd}, grasshopper or a significantly larger animal like a cow. In between come the chickens which are a major source of high-protein and low-fat food \cite{FAO}.

In recent years, there have been unprecedented technological advances in sensor technology, and sensors have become more affordable than ever. Consequently, sensor-driven data collection is increasingly becoming an attractive and practical option for researchers around the globe. Such collected sensor data is typically extracted in the form of time series data; which can be investigated with data mining techniques to summarize the behaviors of the animals \cite{abdoli2018time}.

Time series data is widely used in many domains and is of significant interest in data mining. In recent years researchers have proposed various algorithms for the efficient processing of time series datasets. In this study, we use on-animal sensors to quantify specific behaviors performed by chickens. These behaviors, e.g. preening and dustbathing, are known or suspected to correlate with animal well-being.

\section{How the Data Grows Big?}

Big Data refers to datasets which are big in size and volume; consequently, such datasets are challenging to process and analyze with traditional techniques and software in a timely manner. Datasets are growing rapidly because of the advances in the sensor technologies which enables cheap and mass collection of data. Big data size might range from a few terabytes to zettabytes of data \cite{everts2016information}.

Big Data applications encompass different fields from computer systems and networks \cite{wang2021balancing}\cite{karimi2019border} to smart healthcare/clinical devices \cite{abdoli2018stationary}\cite{abdoli2015field}\cite{abdoli2015general}\cite{abdoli2020cost} and retail industry. However, in order to interpret and analyze the data it should be prepared and cleansed, so that the data can be used by downstream algorithms such as clustering and classification algorithms. In this study, we discuss data management techniques for on-animal sensors to quantify specific behaviors performed by chickens.

\section{Big Data in Poultry Science}

Given the ever-increasing population of the world, the demand for such food sources has been steadily growing. According to Food and Agriculture Organization of the United Nations (FAO), Poultry is the world’s primary source of animal protein. Between 2000 and 2030, per capita demand for poultry meat is projected to increase by 271 \% in South Asia, 116 \% in Eastern Europe and Central Asia, 97 \% in the Middle East and North Africa and 91 \% in East Asia and the Pacific \cite{FAO}. In developed countries, there are growing concerns about the ethical treatment of these animals; among which are housing conditions and how the animals are managed and treated.

Arthropod ectoparasites reside on the surface of the body of chickens, causing stress to the host, and potentially spreading to nearby chickens or other animal hosts \cite{murillo2017review}. Many of these ectoparasites, such as the northern fowl mite, adversely affect productivity (e.g. laying eggs) and health of the chickens. They may also impact poultry behavior and welfare. Understanding how the chicken behave (i.e. timing and frequency of chicken behaviors) can help producers determine behavior abnormalities due to infestations and deploy preventive and corrective control methods. We are not proposing that all chickens be monitored, that is clearly unfeasible. Our system is designed as a tool to allow researchers to assess the effects of various conditions on chicken health, and then use the lessons learned on the entire brood.\\

\begin{figure}[h]
\centering
\includegraphics[width=.6\textwidth]{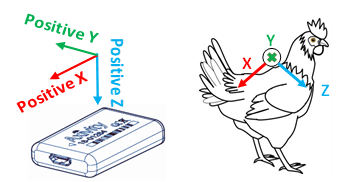}
\caption{(left) Axivity AX3 axis alignment (right) positioning of AX3 sensor on the back of chicken.}~\label{fig:Axivity}
\end{figure}

\section{Collecting Chicken Data}

All chickens were housed and cared for in accordance with UC Riverside Institutional Animal Care and Use Protocol. Data is collected from chickens by placing the sensor on bird’s back. The sensor is placed on back of the bird to allow for high-quality recording of various types of typical chicken behaviors, with the minimum interference and discomfort. The Axivity AX3 sensor used in our study, weighs about 11 grams and is configured with 100 Hz sampling frequency and +/- 8g sensitivity which allows for two weeks of continuous data collection with the battery fully charged. Figure \ref{fig:Axivity} (left) shows the orientation of Axivity AX3 sensor (right) shows placement of sensor on the back of the chicken.

\begin{figure}[h]
\centering
\includegraphics[width=\textwidth]{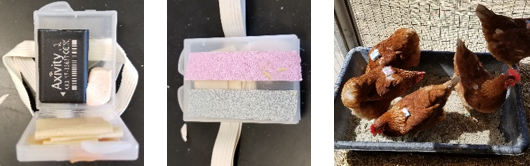}
\caption{(left) Axivity AX3 sensor secured inside a plastic backpack with a rubber band wrist (center) The backpack ready to be mounted on a chicken with color markers for facilitated visual recognition (right) Chickens wearing backpacks on their back. Photos courtesy of Amy C. Murillo.}~\label{fig:Axivity_Real}
    \vspace{-1cm}
\end{figure}

Figure \ref{fig:Axivity_Real} (left) shows the Axivity AX3 sensor secured inside a plastic backpack with a rubber band to allow placing the backpacks on the back of chickens (center) The backpack ready to mounted on the chicken with color markers for facilitated human recognition (right) shows placement of sensors on the back of the chickens. We used USB hubs for mass charging of the sensors; which allowed for simultaneous charging of multiple sensors towards saving human time and effort. After the sensors were fully charged the team member Amy C. Murillo connected each sensor to a desktop computer and utilized the AX3 GUI software to setup the sensors to start collecting data at some predefined date and time.\\

\begin{figure}[t]
\centering
\includegraphics[scale=0.35]{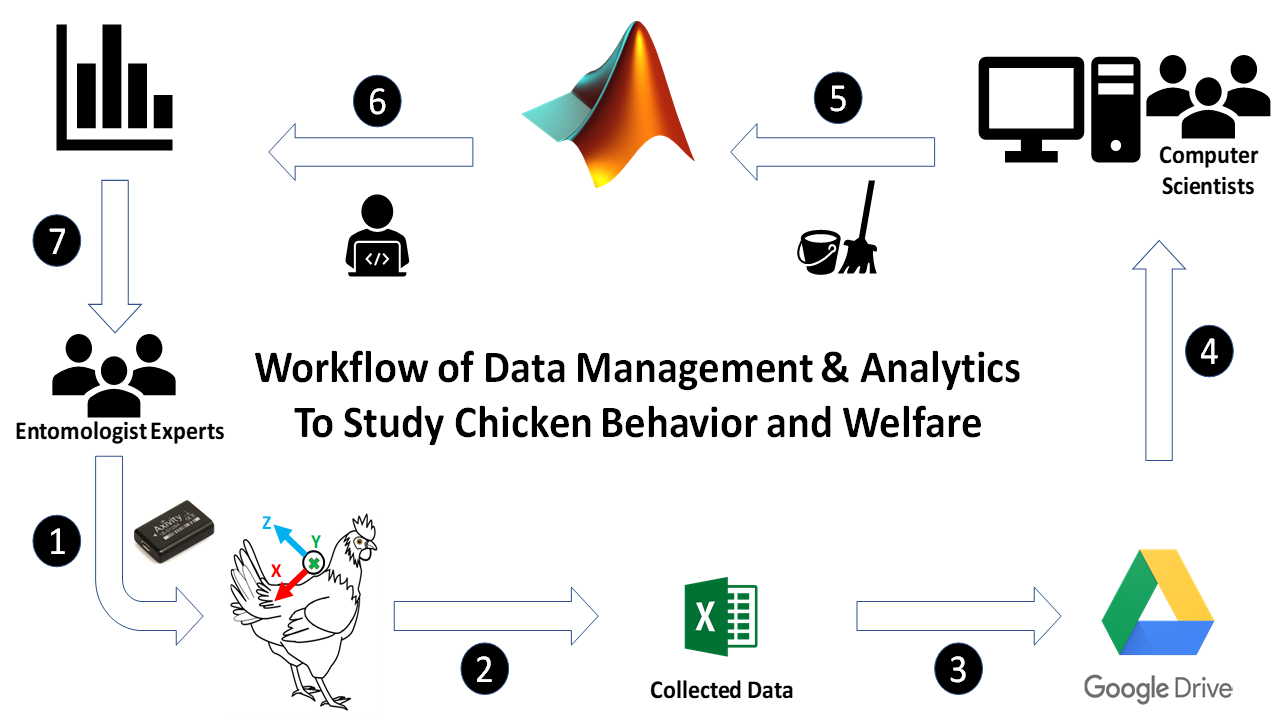}
\caption{The workflow to study chicken behavior and welfare.}~\label{fig:Flow}
    \vspace{-1cm}
\end{figure}

The overview of the processes to study chicken behavior and welfare is shown in Figure \ref{fig:Flow}. This workflow repeated for each of the readings. The description of the individual processes is provided in the following:\\

\begin{figure}[t]
\centering
\includegraphics[scale=0.4]{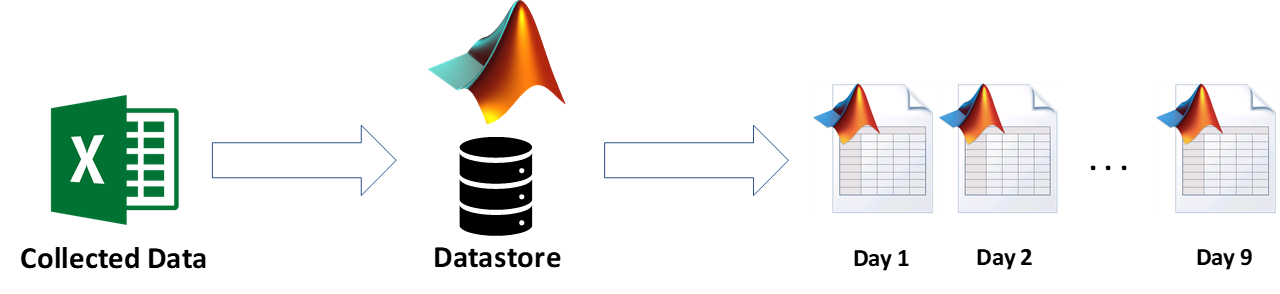}
\caption{The slicer algorithm leverages the datastore in MATLAB and breaks the CSV file into individual days.}~\label{fig:Chunks}
    \vspace{-1cm}
\end{figure}

\begin{enumerate}
    \item The entomologist experts setup the sensors and make sure the sensors are fully charged and then the backpacks are mounted on the chickens.
    
    \item The sensors collect data for 7-9 full days per reading. The data is stored in the CSV file format on the sensor’s internal memory.
    
    \item The entomologist experts pickup the sensors from chickens and upload the chicken data file (CSV file format) onto the Google Drive for facilitated accessibility and availability of data for all team members.
    
    \item Following successful upload of data onto the Google Drive.
    
    \item Next, computer scientists proceed with the developed algorithm for cleansing and preparation of chicken data (as in Figure \ref{fig:Chunks}) for further downstream processing (e.g. classification, clustering and etc.).
    
    \item Given the cleansed and well-formatted data, the computer scientists can proceed with classification of the chicken behaviors. The classification results demonstrate the timing and frequency of various chicken behaviors throughout the 24-hours.

    \item The classification results (i.e. behavior counts and frequency) are provided to entomologist experts so they can look for anomalies and irregularities in the chicken behavior and welfare.
\end{enumerate}

\section{Algorithm For Managing Chicken Big Data}

This section presents the algorithm developed for pre-processing and cleansing of data collected from sensors mounted on chickens. The chicken data downloaded off the sensors were into the format of single large CSV files per sensor holding data for the entire reading period (i.e. 7-9 days). The files were as big as 3-4 GB per file (several terabytes of chicken data); the large file sizes made them challenging to work with and process in MATLAB. Also, the entomologist experts preferred to work on individual days basis. So, an algorithm was developed to take in each CSV file and break it down into individual days. Also, the CSV file format was pretty space-consuming, so the developed algorithm stored the individual days as MATLAB friendly MAT files. The most challenging part was to read the CSV files into memory and slice it into individual days; however, as the CSV files were large in volume, they would cause problems.

In order to overcome the large size of CSV files we utilized the Datastore concept in MATLAB. The datastore essentially is a repository for collections of data that are too large to fit in memory. A datastore allows you to read and process data stored in multiple files on a disk, a remote location, or a database as a single entity. If the data is too large to fit in memory, you can manage incremental import of data for further processing \cite{everts2016information}.

\begin{figure}[!b]
\centering
\includegraphics[width=.8\textwidth]{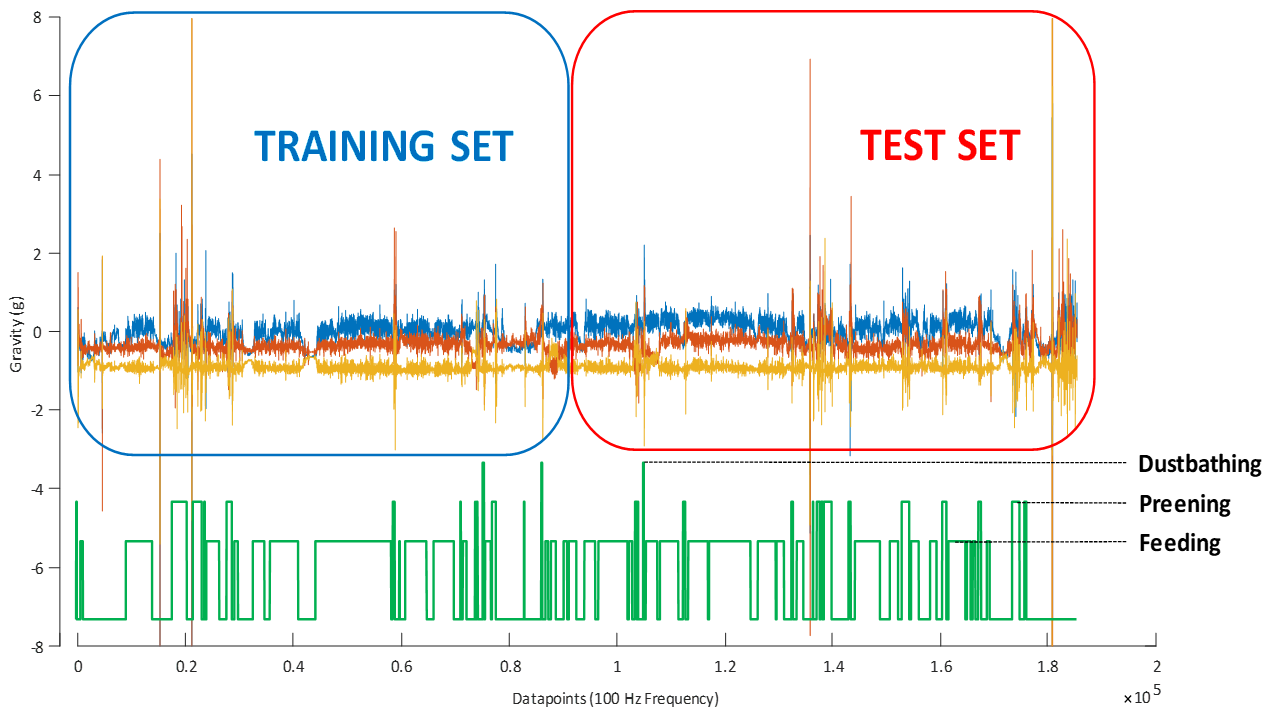}
\caption{Three-dimensional chicken time series (the top/\textcolor{blue}{blue} time series is X-axis; middle/\textcolor{red}{red} time series is Y-axis and bottom/\textcolor{orange}{yellow} time series is Z-axis time series). The green lines represent annotations of observed chicken behaviors captured on video.}~\label{fig:icmla_day}
\end{figure}

\section{Chicken Behavior Classification}

Given the data being cleansed and prepared, the computer scientist experts may proceed with classification of chicken behaviors for each full day. The computer scientists then provide the behavior counts to the entomologist experts so they can comment about potential anomalies in the behavior count and intervene with preventive and corrective actions. The classification of chicken behaviors throughout the day has been discussed in \cite{abdoli2018time}; in which computer scientists extracted a dictionary of well-preserved chicken behaviors (i.e. pecking, preening and dustbathing) from a short video annotated chicken dataset, as shown in Figure \ref{fig:icmla_day}.

\begin{figure}[!b]
\centering
\includegraphics[width=.8\textwidth]{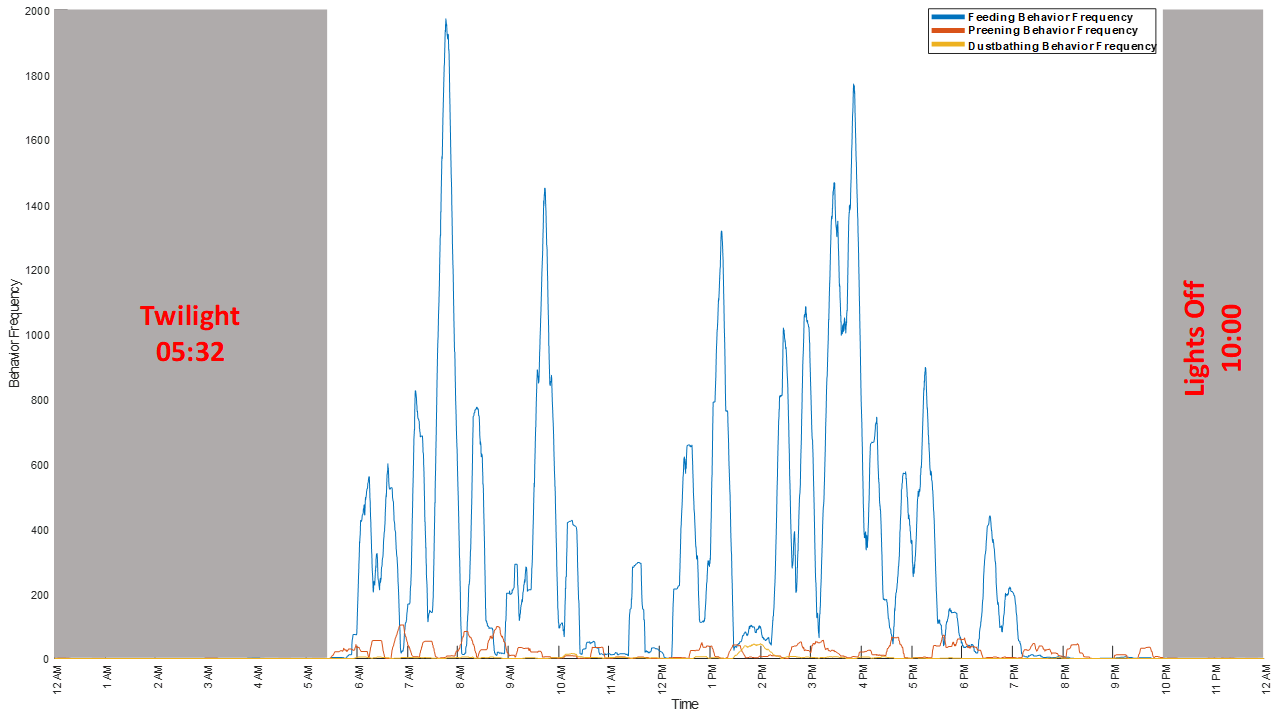}
\caption{Temporal frequency of chicken behaviors (\textcolor{blue}{feeding/pecking}, \textcolor{red}{preening} and \textcolor{orange}{dustbating}) for the twenty-four hour chicken dataset.}~\label{fig:icmla_circadian}
\end{figure}

Then, used the dictionary of behaviors on the unseen datasets from other birds. Following the classification of bird behaviors for every individual day the count and timing of behaviors are reported by the classification algorithm. This helps the entomologist experts to analyze the behavior count and timing and proceed with potential actions. Figure \ref{fig:icmla_circadian} shows the circadian plot on timing and frequency of a full-day dataset.

\section{Conclusions}

In this chapter, we discussed processes involved with big data management. The suggested processes can be applied to any big dataset; while, this study applied the cleansing and pre-processing algorithm to the chicken datasets to classify and quantify chicken behaviors towards studying chicken welfare. Further, the chapter briefly addressed classification and visualization of chicken behaviors in a more intuitive way to the audience.
\chapter{Shape-based Classification: Time Series Classification to Improve Poultry Welfare}

\newpage

\section{Introduction}

Domesticated birds, e.g. chickens, are a major source of food for humans; people have been keeping birds for their eggs (laying chickens) and for their meat (broiler chickens) for thousands of years \cite{tixier2011chicken}. Poultry farms are a major source of high-protein and low-fat food.

Given the ever-increasing population of the world, the demand for such food sources has been steadily growing. According to Food and Agriculture Organization of the United Nations (FAO), poultry meat consumption around the world has climbed from 11 kg/person in 2000 to 14.1 kg/person in 2011 \cite{ThePoultrySite}; and it is predicted to continue for the foreseeable future \cite{FAO}. In developed countries, there are growing concerns about the ethical treatment of these animals; among which are housing conditions and how the animals are managed and treated \cite{blatchford2017animal}.

Ectoparasites are a group of arthropods that reside on the surface of the body of chickens, causing stress to the host, and potentially spreading to nearby chickens or other animal hosts \cite{murillo2020parasitic}. Many of these ectoparasites, such as the northern fowl mite, adversely affect productivity (e.g. laying eggs) and health of the chickens [14]. They may also impact poultry behavior and welfare \cite{brown1974effect}. Understanding how ectoparasites affect chicken behavior can help producers determine when flocks are infested, to better deploy control methods \cite{murillo2016sulfur}. Traditional behavior studies have relied on direct or video observation of subjects. However, this is time consuming, error-prone and subjective.

\begin{figure}[h]
\centering
\includegraphics[width=0.8\textwidth]{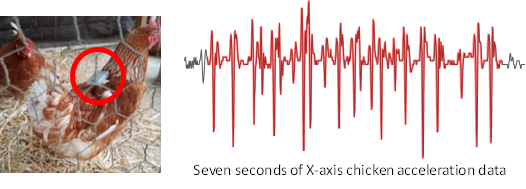}
\caption{(left) A chicken with an Axivity AX3 accelerometer worn inside a ‘backpack’ on the back of chicken (right) A seven-second snippet of chicken time series data collected from the accelerometer.}~\label{fig:fig_1}
\end{figure}

We argue that the use of on-animal sensors can help to increase richness and density of observations \cite{siegford2016assessing}. Additionally, sensors can be used to greatly increase the number of individuals that can be tracked, while also expanding the tracking period, in some cases to 24/7 monitoring. In recent years, there have been enormous technological advances in sensor technology, and consequently sensor prices have decreased dramatically; this in turn has made sensor-driven data collection a practical option. Recently, there have been various studies on using sensors for collecting data in the context of livestock and poultry \cite{smythe2015behavioral}\cite{okada2010avian}, even flying insects have not escaped the efforts at digitally sensed monitoring \cite{batista2011sigkdd}.

Such collected sensor data is typically extracted in the form of time series data; which can be examined offline with data mining techniques to summarize the behaviors of the animals under study. Time series data is widely used in many domains and is of significant interest in data mining \cite{gao2017iterative}\cite{giusti2016improved}\cite{lee2017anomaly}\cite{da2017clustering}\cite{parmezan2015study}\cite{imani2021multi}\cite{imanitime2cluster}. In recent years researchers have proposed various algorithms for the efficient processing of time series datasets \cite{yeh2016matrix}\cite{gharghabi2017matrix}\cite{dau2017matrix}\cite{imani2018matrix}, even in case of noisy data \cite{gharghabi2018matrix}. In this chapter, we use on-animal sensors to quantify specific behaviors performed by chickens. These behaviors, e.g. dustbathing, are known or suspected to correlate with animal well-being.

\section{Studying Chicken Behaviors}

Before moving on, we will take the time to ward off a possible misunderstanding. We are not proposing that all chickens be monitored, that is clearly unfeasible. Our system is designed as a tool to allow researchers to assess the effects of various conditions on chicken health, and then use the lessons learned on the entire brood. Our particular motivating example initiated with a study at UC Riverside to assess the effects of ectoparasites on chickens. However, our methods can be used to determine the effects of any change in the bird’s environment or diet. Our only assumption is that the change will manifest in changes in the frequency or timing of the bird’s behavior(s).

\subsection{Why is studying chicken behaviors at scale hard?}

There are hundreds of studies on quantifying human body behaviors with sensors. Such studies typically involve finding discrete well-defined classes of behaviors, and then monitoring data for future occurrences of behaviors. One example is “step-counting” to measure compliance with a suggested exercise routine. However, the task of studying chicken behaviors is more difficult for the following reasons:

\begin{itemize}
    \item In case of humans, the sensors can be easily placed on the extremities of the limbs (i.e. smart-shoes or smartwatches); However, the placement of sensors on chickens has been primarily restricted to the back of the birds, see Figure 1 (left), due to sensor limitations and the welfare of animal. This provides only coarse information about the bird’s behaviors.\\
    
    \item The variability of human behaviors is well-studied, and it is understood what fraction can be attributed to individual personality, mood and so forth, versus variability in sensor placement \cite{doppler2009variability}. More importantly, it is known how to account for this variability.  However, it is less clear how much variability exists in birds and how we can best account for it.
    
    \item Creating a dictionary for a human subject is relatively straight-forward. During an explicit training session, behaviors of interest can be acted out in a fixed order for a fixed duration of time. Perhaps the most studied human motion time series dataset is gun-point \cite{ding2008querying}. When recording that dataset, the actor’s behaviors were cued by a metronome. Chickens are clearly not as cooperative, and many hours/days of video recordings must be analyzed to create a behavior dictionary. Moreover, it is difficult, even for an experienced avian ethologist, to define precisely where a behavior begins and ends, thus we must be able to work with “weakly labeled” data.
\end{itemize}

In this chapter, we introduce a novel dictionary learning algorithm which can take weakly labeled data in the format “there are a few pecks somewhere in this time period,” together with some mild constraints “a preening behavior probably lasts between 0.3 and 1.5 seconds” and automatically construct a dictionary of behaviors. As we shall show, this dictionary can then be used to classify unlabeled archives of bird behavior.

\section{Related Work and Background}

In recent works \cite{walton2018evaluation}\cite{barwick2018categorising}, sensors were used for classification of sheep behaviors; with mounted sensors on ear/collar or leg of the sheep. Similar work has been performed for domestic bovines \cite{smythe2015behavioral}, and for various kinds of wild animals. In addition, there has been some work on poultry behaviors using both sensors and human monitoring \cite{moreau2009use}\cite{banerjee2012remote}. However, this work is complementary to our efforts. They use statistical features of the accelerometer (mean, entropy, etc.) to quantify periods of general behaviors, such as sleep, stand, walk, etc. \cite{banerjee2012remote}.  In contrast, we use the shape of the time series to precisely annotate very specific and dynamic behaviors, such as individual instances of a single peck. By analogy with human studies, this is similar to recognizing the difference between when someone having lunch versus recognizing each individual bite. Both types of information can be useful for various tasks. We begin by providing definitions and notation to be used throughout the chapter.

\section{Definitions and Notation}

\textbf{Definition 1:} A time series $T$ is a sequence of real-valued numbers $t_i: T = [t_1, t_2, …, t_n]$ where $n$ is the length of $T$.

We are typically not interested in the global properties of time series, but in the similarity between local subsequences:

\textbf{Definition 2:} A subsequence $T_{i,m}$ of a time series $T$ is a continuous subset of the values from $T$ of length $m$ starting at position $i$. $T_{i,m} = [t_i, t_{i+1}, …, t_{i+m-1}]$ where $1 \leq i \leq n – m + 1$

We can take any subsequences and calculate its distance to all subsequences in a time series. An ordered vector of such distances is called a distance profile:

\newpage
\textbf{Definition 3:} A distance profile $D$ is a vector of Euclidean distances between a given query and each subsequence in the time series.

\begin{figure}[h]
\centering
\includegraphics[width=0.7\textwidth]{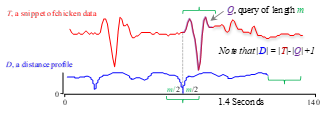}
\caption{Distance profile D obtained by searching for query Q (here the query is extracted from the time series itself) in time series T.}~\label{fig:fig_2}
\end{figure}

Figure \ref{fig:fig_2} illustrates calculating distance profile (D). It is assumed that the distance is measured using the Euclidean distance between Z-normalized subsequences. In Figure \ref{fig:fig_2}, the query Q is extracted from time series T. As can be seen, distance profile has low values at the location of subsequences which are highly similar to the query Q. In case the query Q is taken from the time series itself, then the value for the distance profile at the location of query should be zero, and close to zero just before and just after. To avoid such trivial matches an exclusion zone with the length of ($m/2$) is placed to the left and right of the query location. The distance profile can be computed very efficiently using the MASS algorithm \cite{mueen2017fastest}.

\section{Dictionary of Chicken Behaviors}

\subsection{Collecting and Preparing Chicken Data}

All chickens were housed and cared for in accordance with UC Riverside Institutional Animal Care and Use Protocol A-20150009. Data is collected from chickens by placing the sensor on bird’s back, as shown in Figure \ref{fig:fig_1} (left). The sensor is placed on back of the bird to allow for high-quality recording of various types of typical chicken behaviors, with the minimum interference and discomfort. The Axivity AX3 sensor used in our study, weighs about 11 grams and is configured with 100 Hz sampling frequency and +/- 8 g sensitivity which allows for two weeks of continuous data collection with the battery fully charged.

From literature reviews , and conversations with poultry experts, we expect that the following behaviors correlate with poultry health:

\begin{itemize}
    \item \textbf{Feeding/pecking:} bringing the beak to the ground; striking at the ground.
    \item \textbf{Preening:} preening/grooming of the feathers by the beak; feathers may be drawn or nibbled by the beak \cite{daigle2014moving}.
    \item \textbf{Dustbathing:} bird is in a sitting or lying position with feathers raised in a vertical wing-shake \cite{daigle2014moving}.
\end{itemize}

For some fraction of the time series data collected, a video camera also recorded the chicken activity ($\sim$ 30 minutes). This provides ground truth to act as training data. The sensor data was carefully annotated \cite{ELAN}, based on the video-recorded chicken activities by team member Amy C. Murillo; it must be noted that even the most careful human labeling of chicken behaviors can contain errors, especially false negatives.

More importantly, due to technical limitations it is difficult to synchronize the two data sources to anything less than one-second variable lag, which is a long time relative to a chicken peck ($\sim$ 1/4 of a second). Thus, as shown in Figure 4, the annotations take the form of a categorical vector (shown in green) that indicate that in the corresponding region one or more examples of the corresponding behavior were observed. Such data is often called “weakly labeled” data.

To reiterate, inspecting this video is very time consuming for a technician; we do not propose to inspect all data this way, this is simply a one-time ground truthing operation performed for a tiny fraction of the data collected.

\subsection{Creating a Dictionary of Behaviors}

Given that we have training/test data in the format shown in Figure 4, we are in a position to attempt to automatically construct a dictionary of behaviors or query-templates.

A dictionary is a list of query-templates (behaviors) in the form of {s1, s2, …, si}; each query has a class (s1.class), a threshold value (s1.threshold) and axis (s1.axis) properties, along with the query data points. 

In principle, a single behavior could have two or more possible instantiations; just like the number four has two written versions, closed ‘4’ and open ‘’, which are semantically identical. Such a dictionary is called a polymorphic dictionary. Given our observations of chicken behaviors, in this chapter we assume that there is a single way to perform a behavior. However, generalizing the code to a polymorphic dictionary is trivial.

\subsection{Algorithms for Building the Chicken Behavior Dictionary}

Our algorithm for building a dictionary of chicken behaviors works by searching within annotated regions for highly conserved sequences (i.e. motifs). To give an intuition for this, consider the analogue problem in the discrete string space. Imagine we are given this data snippet, and we are told that the green region is a weak label for pecking.

Further suppose we are told that the length of the behavior is between 2 and 5 symbols. If we looked for conserved behavior of length 2, we would find that ‘be’ happens six times. However, three of those occurrences are outside the annotated region, so this cannot be a good predictor of the class. If we looked for conserved behavior of length 3, we would find that ‘bes’ happens three times, and all of the occurrences happen within the labeled region. This seems like a better predictor of the class. However, note that if we expand to conserved pattern of length 4, ‘beso’ also happens three times within the labeled region. Since we might expect that the longer pattern is more specific to the class, we prefer it. Note that if we continue the search to patterns of length five, there are no highly conserved patterns of this length.

In addition to time series T and the annotation labels Label, the algorithm takes a range of lengths Len (equivalent to the values 2 to 5 in the example above). Algorithm \ref{ICMLA_ALG_1} presents the pseudo-code for building a dictionary of behaviors. In Line 1 the range of query-template lengths to be tested is specified. Line 2 iterates over the annotated regions. In Line 3, a sliding window is initiated inside the selected region from Line 2 with the specified length from Line 1. The selected query-template, time series, label data are provided to the nearest neighbor similarity search subroutine algorithm. The output result for the selected query-template is added to the current list of query-templates (QueryList) in Line 6.


\begin{algorithm}
\textbf{Algorithm for Building Dictionary}
\label{ICMLA_ALG_1}
\begin{figure}[h]
\centering
\includegraphics[width=\textwidth]{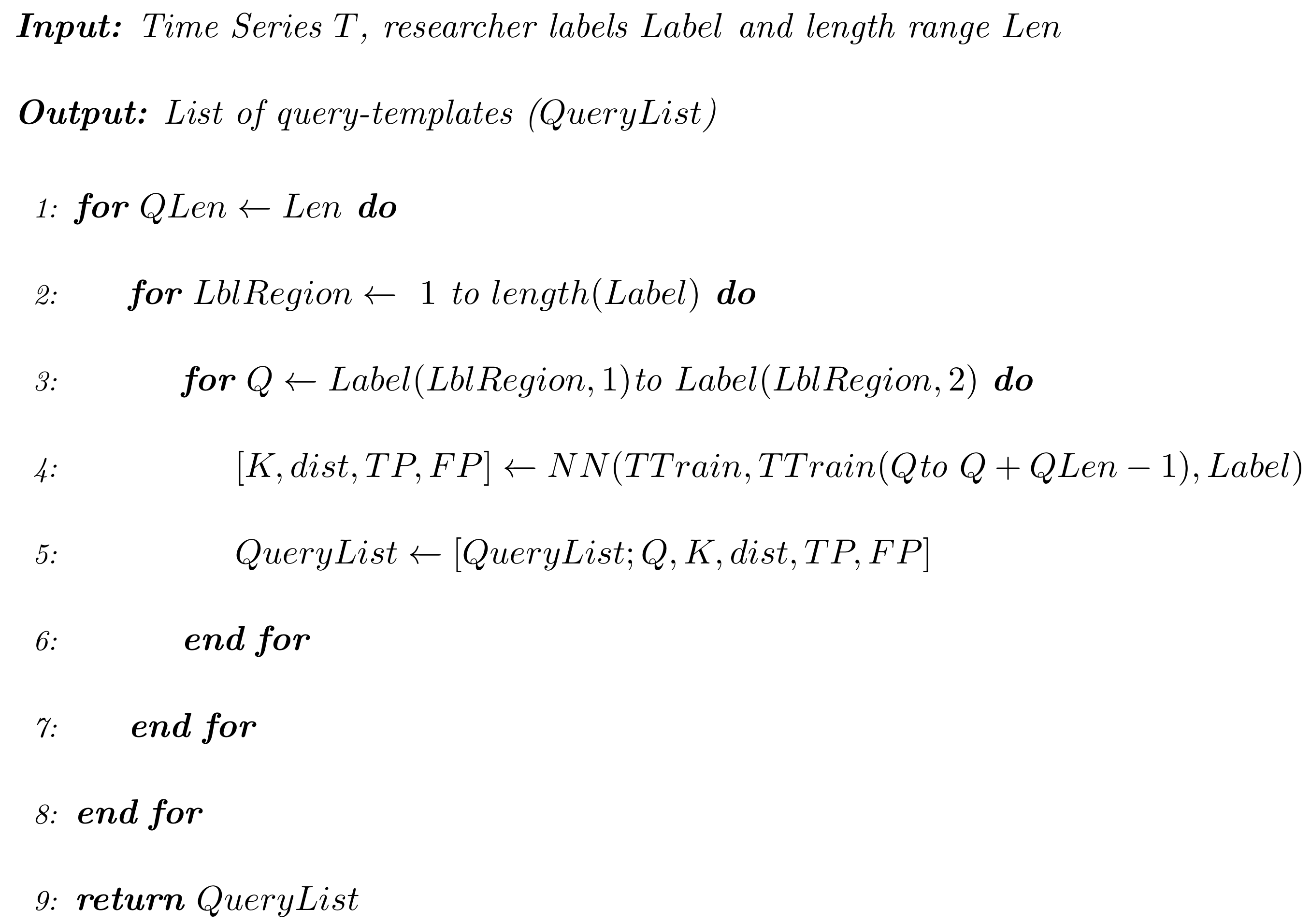}
\end{figure}
\end{algorithm}

\newpage

Algorithm 2 is a subroutine, which given the candidate query-template, time series and label data calculates the similarity between the query-template and all subsequence in the time series.


\begin{algorithm}
\textbf{Nearest Neighbor Similarity Search Algorithm}
\label{ICMLA_ALG_2}
\begin{figure}[h]
\centering
\includegraphics[width=\textwidth]{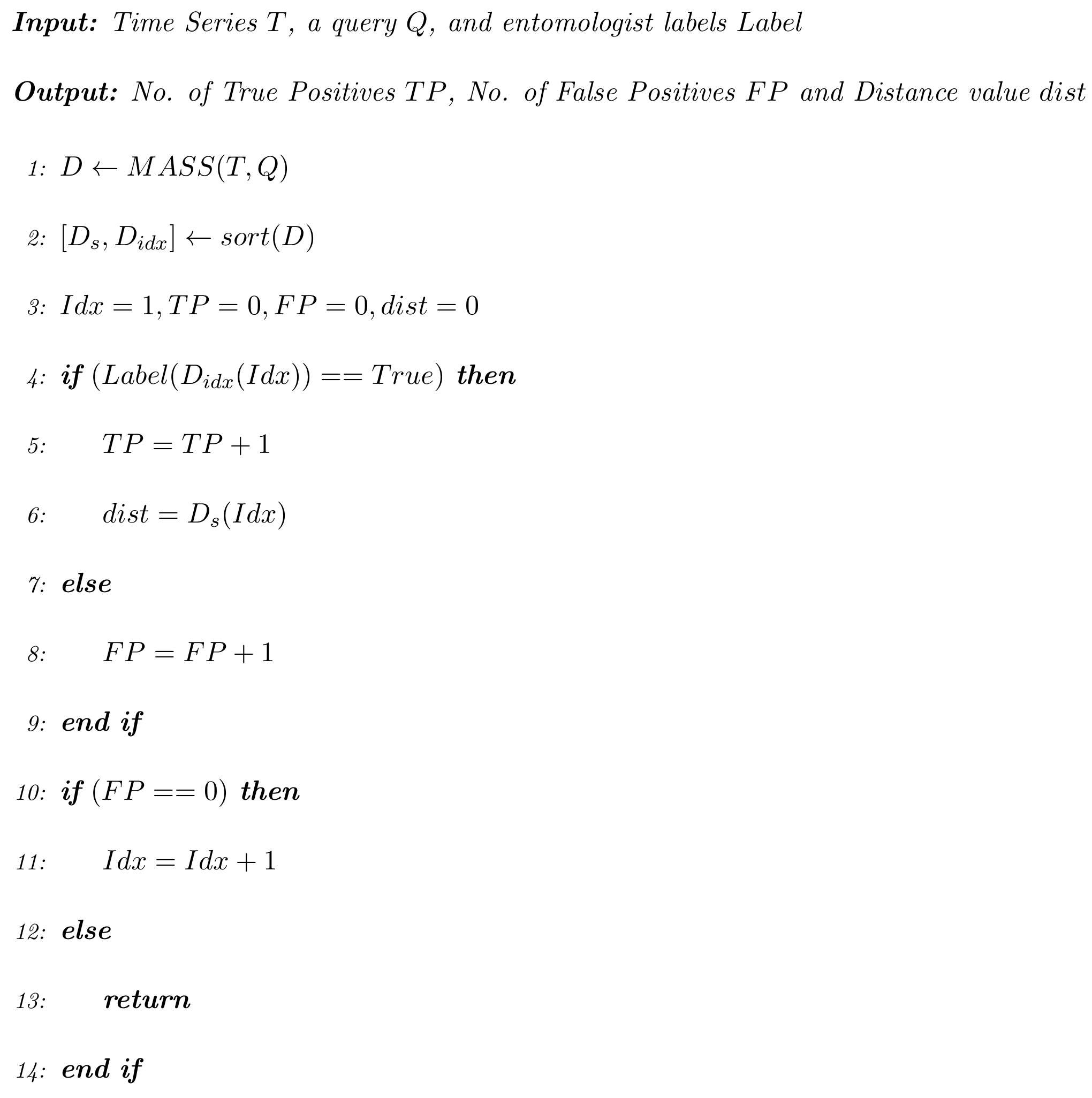}
\end{figure}
\end{algorithm}

In Line 1, the candidate query-template and time series data are passed to the MASS algorithm. MASS computes the similarity between query-template (Q) and every subsequence in time series (T); and returns a distance profile D \cite{mueen2017fastest}. The distance profile D is sorted in ascending order (Ds) and the indices of sorted values are stored in $D_{idx}$ in Line 2.

In Lines 4 - 9, at each iteration, the algorithm takes a value from (Didx(Idx)) and examines the corresponding value of Label(Didx(Idx)) to see if it is marked as a targeted behavior. If it is marked as a behavior of interest, then the subsequence is a true positive (TP) match and TP is increased by one (Line 5), otherwise, it is treated as a false positive (FP) match and FP is increased by one (Line 8). We are interested in candidate query-templates which yield the highest number of true positive matches with no false positive matches. Therefore, in Lines 10 - 14, we continue the search as long as no false positive (FP) match is observed. It is worth noting that the distance (dist) mentioned in Line 6 of Algorithm \ref{ICMLA_ALG_2} is the same as query-template threshold value in the behavior dictionary. The threshold value serves as a measure of similarity when searching in unlabeled data streams. In the case, a subsequence in the unlabeled data has a similarity value with the query-template lower than the threshold it is classified as a matching subsequence.

\subsection{How to use the Chicken Behavior Dictionary?}

When monitoring the stream of time series data, we look for specific behaviors to count, classify, and time-stamp. The stream of data is processed through a sliding window. In case the similarity threshold is met for some query-template (behavior) in the dictionary, then the sequence is matched and time-stamped as an instance of that behavior.

\section{Empirical Evaluation}

To ensure that our experiments are reproducible, we have built a supporting website; which contains all data, code and raw spreadsheets for the results, in addition to many experiments that are omitted here for brevity.

We provide evaluation results for the feeding/pecking, preening, and dustbathing behaviors. The original dataset is split into mutually exclusive training and test datasets, as illustrated in Figure 4. The training dataset is used for building the dictionary of behaviors, while the test dataset is solely used for out-of-sample evaluation. As Figure 4 reminds us, the data is weakly labeled, meaning that every annotated region contains one or more of the specified behavior. In addition, there are almost certainly instances of the behavior outside the annotated regions which the annotator failed to label, perhaps because the chicken in question was occluded in the video. However, we believe that such false negatives are rare enough to be ignored. In addition, we do not know about the exact number of individual instances of a behavior inside a region, complicating the evaluation. To address this, we utilize the concept of Multiple Instance Learning (MIL) \cite{amores2013multiple}; which assumes each annotated region as a “bag” containing one or more instances of a behavior.

In this classification model, if at least a single instance of a behavior is matched inside a bag, it is treated as a true positive. However, in case that no instances of the behavior are detected inside the bag, then the entire bag is treated as a false negative. Furthermore, in case an instance of behavior is mismatched inside a bag corresponding to some other behavior, then it is treated as a false positive. Finally, if no mismatch occurs inside a bag of non-relevant behavior, then the entire bag is treated as a true negative.

\subsection{Feeding/Pecking Behavior}

\begin{figure}[h]
\centering
\includegraphics[width=\textwidth]{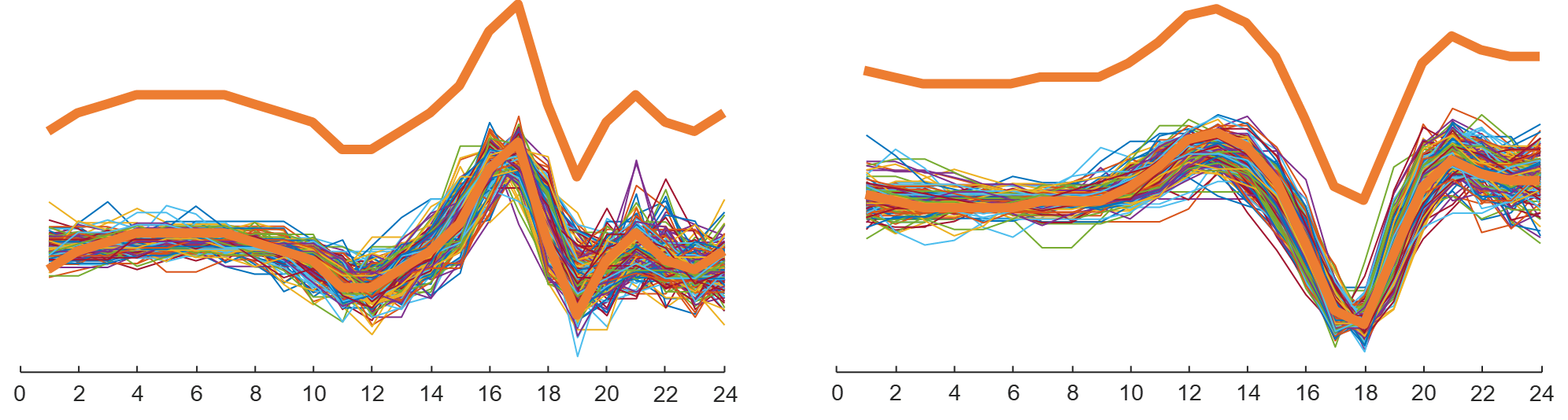}
\caption{Query-template for feeding/pecking behavior, (left) X-axis and (right) Z-axis matching subsequences in the training dataset.}~\label{fig:fig_5}
\end{figure}

Feeding/pecking is perhaps the most familiar behavior in chickens. Figure \ref{fig:fig_5} shows the query-template discovered by our dictionary building algorithm, along with matching subsequences in the training dataset. Recall that subsequences located within regions annotated as containing instances of feeding/pecking behavior are treated as true positives (TP), whereas matches outside of regions annotated for feeding/pecking behavior are treated as false positives (FP).

\begin{figure}[h]
\centering
\includegraphics[width=\textwidth]{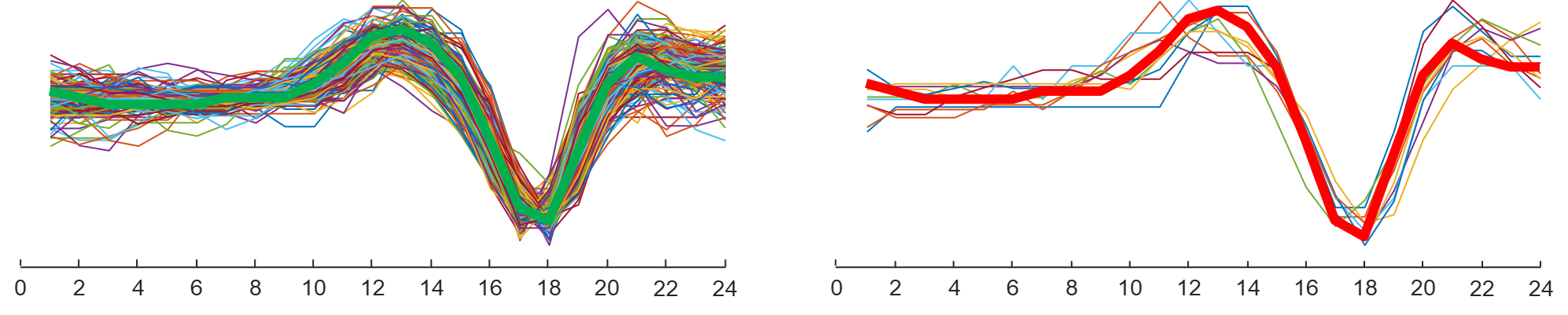}
\caption{The X-axis query-template and matching subsequences for feeding/pecking behavior in the test dataset, (left) true positives (right) false positives. Note that the false positives do appear to actually be true positives that were missed by the human annotator.}~\label{fig:fig_6}
\end{figure}

Figure \ref{fig:fig_6} shows matching subsequences from the test dataset with true positives shown in green and false positives shown in red. The reader will appreciate that the false positives do look at lot like the true positives. As noted above, it is possible (and indeed likely) that they really are true positives that escaped the attention of the human annotator.

Given the positioning of sensor on the back of chicken, as shown in Figure \ref{fig:fig_3} (center) and (right), when the chicken approaches its head to the ground, X-axis and Z-axis are affected. As can be seen in Figure \ref{fig:fig_5} (top), pecking behavior is manifested in the form of a “valley” on X-axis (i.e. negative X-axis acceleration); at the same time as shown in Figure \ref{fig:fig_5} (bottom), the pecking behavior is manifested in the form of a peak on Z-axis (i.e. positive Z-axis acceleration). Further, given that Y-axis (i.e. lateral acceleration towards left or right) is not as influential as X and Z-axis, therefore we omit the Y-axis for this specific behavior. Figure \ref{fig:fig_8} (top) shows the matching subsequences for running the feeding/pecking query-template against the test dataset.

Table \ref{table:icmla_pecking} provides the confusion matrix for the performance of feeding/pecking query-template.

\begin{table}[h]
\caption{Confusion matrix for feeding/pecking behavior}
\begin{center}
\begin{tabular}{|c|c|c|}
\hline
Classes                & Pecking  & Non-Pecking \\ \hline
Pecking        & \textbf{17 (TP)}  & \textbf{7 (FN)}  \\ \hline
Non-Pecking   & \textbf{4 (FP)}  & \textbf{43 (TN)} \\ \hline
\end{tabular}
\end{center}
\label{table:icmla_pecking}
\end{table}

\[
    Accuracy = \frac{TP + TN}{TP + TN + FP + FN} = \textbf{71 \%}
\]

\[
    Precision = \frac{TP}{TP + FP} = \textbf{81 \%}
\]

\[
    Recall = \frac{TP}{TP + FN} = \textbf{85 \%}
\]

Given the results above, our classification model has a 0.71 precision and 0.81 recall in matching instances of the feeding/pecking behavior. Overall, the classifier has 85\% accuracy for the feeding/pecking behavior, which compares very favorably to 70\% default rate (i.e., guessing every observed object as the majority class).

\subsection{Preening Behavior}

\begin{figure}[h]
\centering
\includegraphics[width=0.8\textwidth]{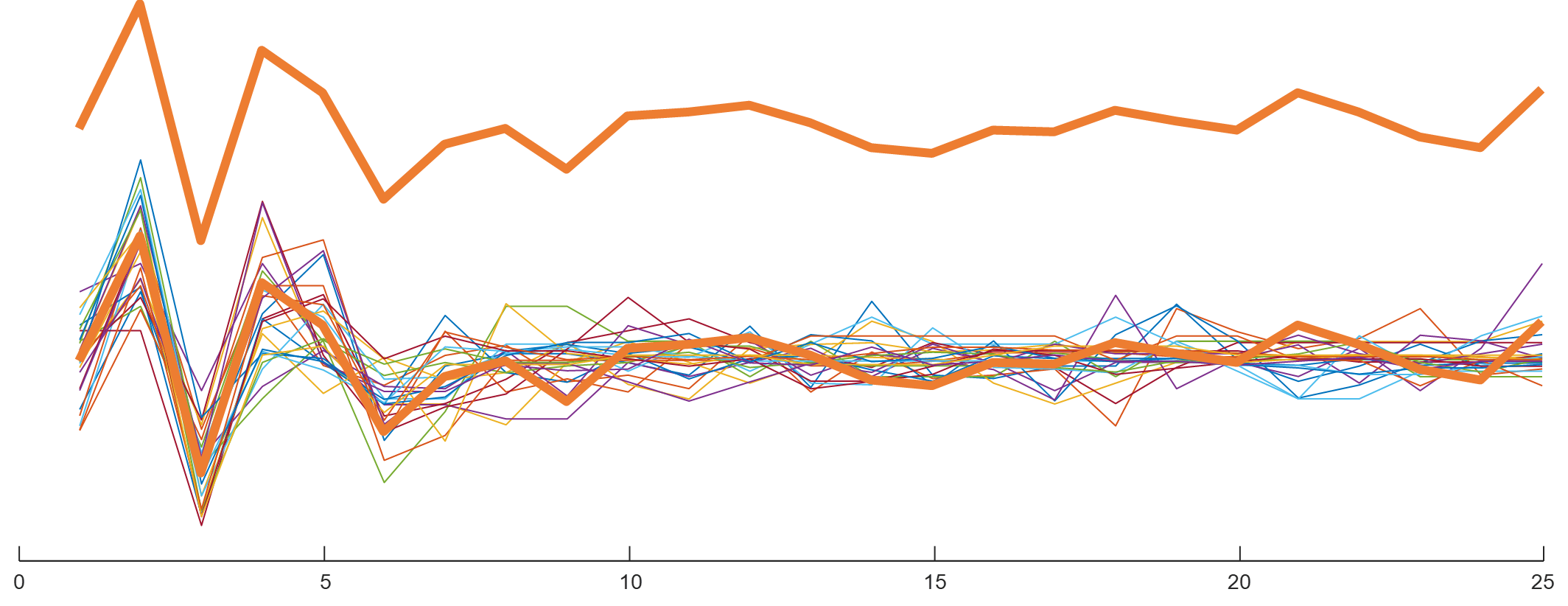}
\caption{The Z-axis query-template corresponding to preening behavior along with matching subsequences in the training dataset.}~\label{fig:fig_7}
\end{figure}

\begin{figure}[t]
\centering
\includegraphics[width=\textwidth]{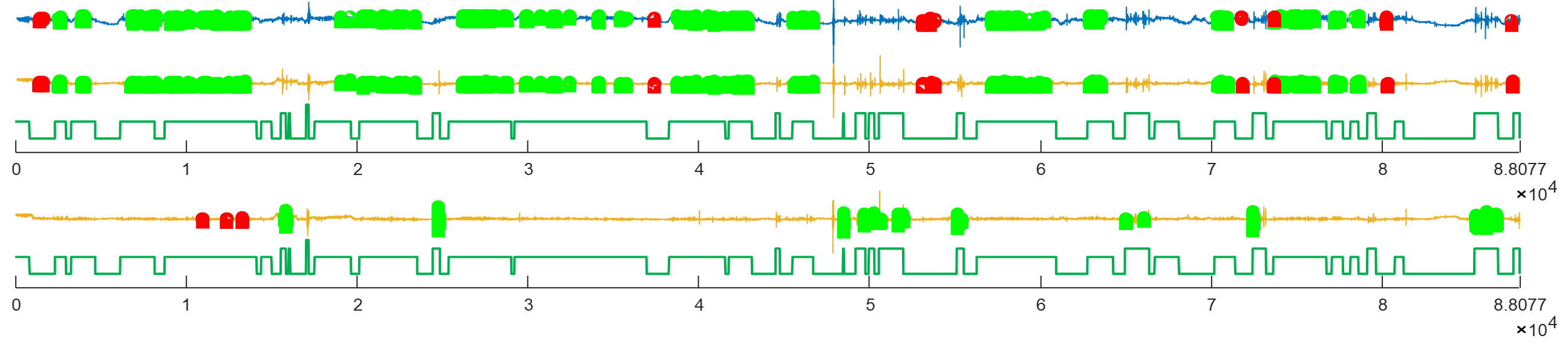}
\caption{Matching subsequences in the test dataset for (top) feeding/pecking behavior on X and Z-axis (bottom) preening behavior only on Z-axis. KEY: green highlights indicate true positives, and red highlights indicate false positives.}~\label{fig:fig_8}
\end{figure}

Preening is the act of cleaning feathers, which is an important part of a chicken’s daily activities. Preening is a grooming behavior that involves the use of the beak to position feathers, interlock feather barbules (informally, “zip up” the feathers) that have become separated, clean plumage, and remove ectoparasites. Figure \ref{fig:fig_7} shows the query-template for the preening behavior and the matching subsequences within the training dataset. Also, Figure \ref{fig:fig_8} (bottom) shows subsequences for running the query-template against a long test dataset.

It is interesting to discuss the nature of the preening behavior before proceeding to evaluation results. One might imagine that a flat region in the data is uninformative, however a flat region directly after the pattern on the Figure \ref{fig:fig_7} is informative. The bird uses its beak to realign the feathers, and as can be seen in Figure \ref{fig:fig_7}, the act of preening starts with moving the head (i.e. positive and negative Z-axis acceleration) and the rest of the pattern is relatively flat which might imply movement of feather through the beak for alignment purposes. Figure \ref{fig:fig_8} (bottom) shows the matching subsequences for running the preening query-template against the test dataset. Table \ref{table:icmla_preening} provides the confusion matrix on the performance of preening query-template.

Given the evaluation results above our classification model has 0.91 precision in matching preening subsequences; and 0.71 recall in matching relevant instances of the preening behavior. Finally, the model has 93\% overall accuracy in matching preening subsequences compared to the 80\% default rate (i.e., guessing every observed object as the majority class).

\begin{table}[h]
\caption{Confusion matrix for preening behavior}
\begin{center}
\begin{tabular}{|c|c|c|}
\hline
Classes                & Preening  & Non-Preening \\ \hline
Preening        & \textbf{10 (TP)}  & \textbf{1 (FN)}  \\ \hline
Non-Preening   & \textbf{4 (FP)}  & \textbf{56 (TN)} \\ \hline
\end{tabular}
\end{center}
\label{table:icmla_preening}
\end{table}

\[
    Accuracy = \frac{TP + TN}{TP + TN + FP + FN} = \textbf{91 \%}
\]

\[
    Precision = \frac{TP}{TP + FP} = \textbf{71 \%}
\]

\[
    Recall = \frac{TP}{TP + FN} = \textbf{93 \%}
\]

\subsection{Dustbathing Behavior}

\begin{figure}[!t]
\centering
\includegraphics[width=0.55\textwidth]{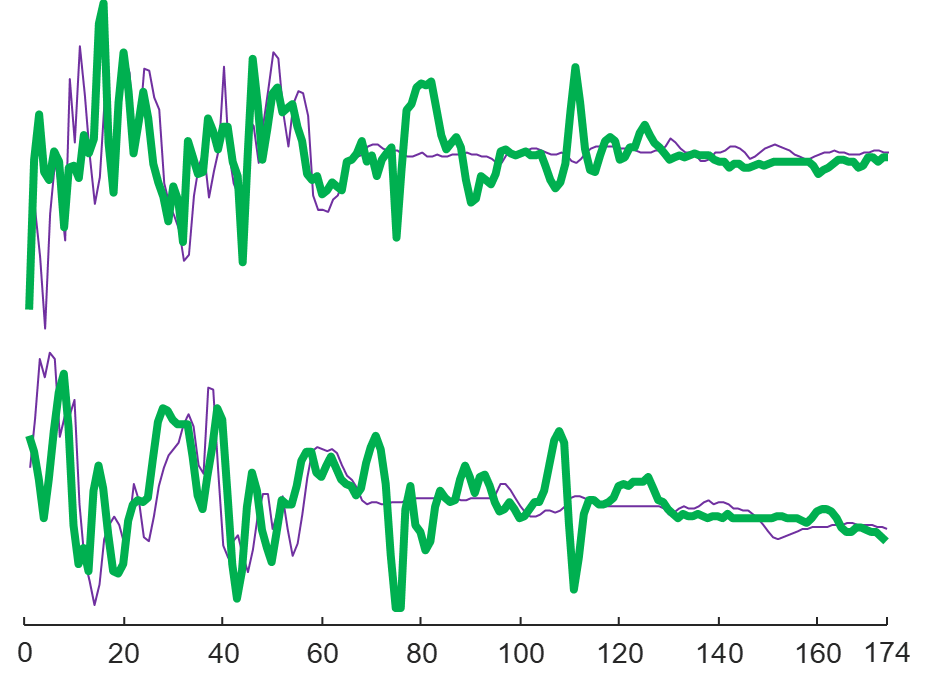}
\caption{Matching subsequence in the test dataset for dustbathing behavior, while this behavior is much rare, our algorithm correctly found the instance. KEY: green highlightings indicate true positives, and red highlightings indicate false positives (we do not have any false positive matches here).
}~\label{fig:fig_9}
\end{figure}

\begin{figure}[t]
\centering
\includegraphics[width=\textwidth]{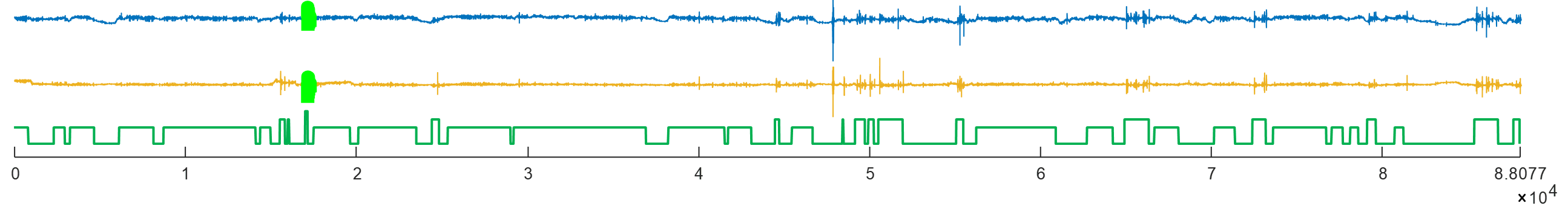}
\caption{Matching subsequence in the test dataset for dustbathing behavior, while this behavior is much rare, our algorithm correctly found the instance. KEY: green highlightings indicate true positives, and red highlightings indicate false positives (we do not have any false positive matches here).}~\label{fig:fig_11}
\vspace{-0.5cm}
\end{figure}

Dustbathing is the act in which the chicken moves around in dust or sand to remove parasites from its feathers. This tends to be the least common activity in chickens. As shown in Figure \ref{fig:fig_9} and Figure \ref{fig:fig_11}, dustbathing tends to be the most difficult behavior to search for, since there were only two instances in the training dataset and only a single instance in the test dataset. Table \ref{table:icmla_dustbathing} presents confusion matrix for the dustbathing behavior.

\begin{table}[!h]
\caption{Confusion matrix for dustbathing behavior}
\begin{center}
\begin{tabular}{|c|c|c|}
\hline
Classes                & Dustbathing  & Non-Dustbathing \\ \hline
Dustbathing        & \textbf{1 (TP)}  & \textbf{0 (FN)}  \\ \hline
Non-Dustbathing   & \textbf{0 (FP)}  & \textbf{70 (TN)} \\ \hline
\end{tabular}
\end{center}
\label{table:icmla_dustbathing}
\end{table}

Given these evaluation results, our model has 1.00 precision in matching dustbathing subsequences and 1.00 recall in matching relevant instances of the dustbathing behavior.

\[
    Accuracy = \frac{TP + TN}{TP + TN + FP + FN} = \textbf{100 \%}
\]

\[
    Precision = \frac{TP}{TP + FP} = \textbf{100 \%}
\]

\[
    Recall = \frac{TP}{TP + FN} = \textbf{100 \%}
\]

Finally, the model has 100\% overall accuracy in matching dustbathing subsequences compared to 99\% default rate (i.e., guessing every observed object as the majority class). Figure \ref{fig:fig_11} shows subsequences for running the dustbathing query-template against the test dataset.

\section{Case Study: An Entire Day With A Chicken}

\begin{figure}[h]
\centering
\includegraphics[width=.9\textwidth]{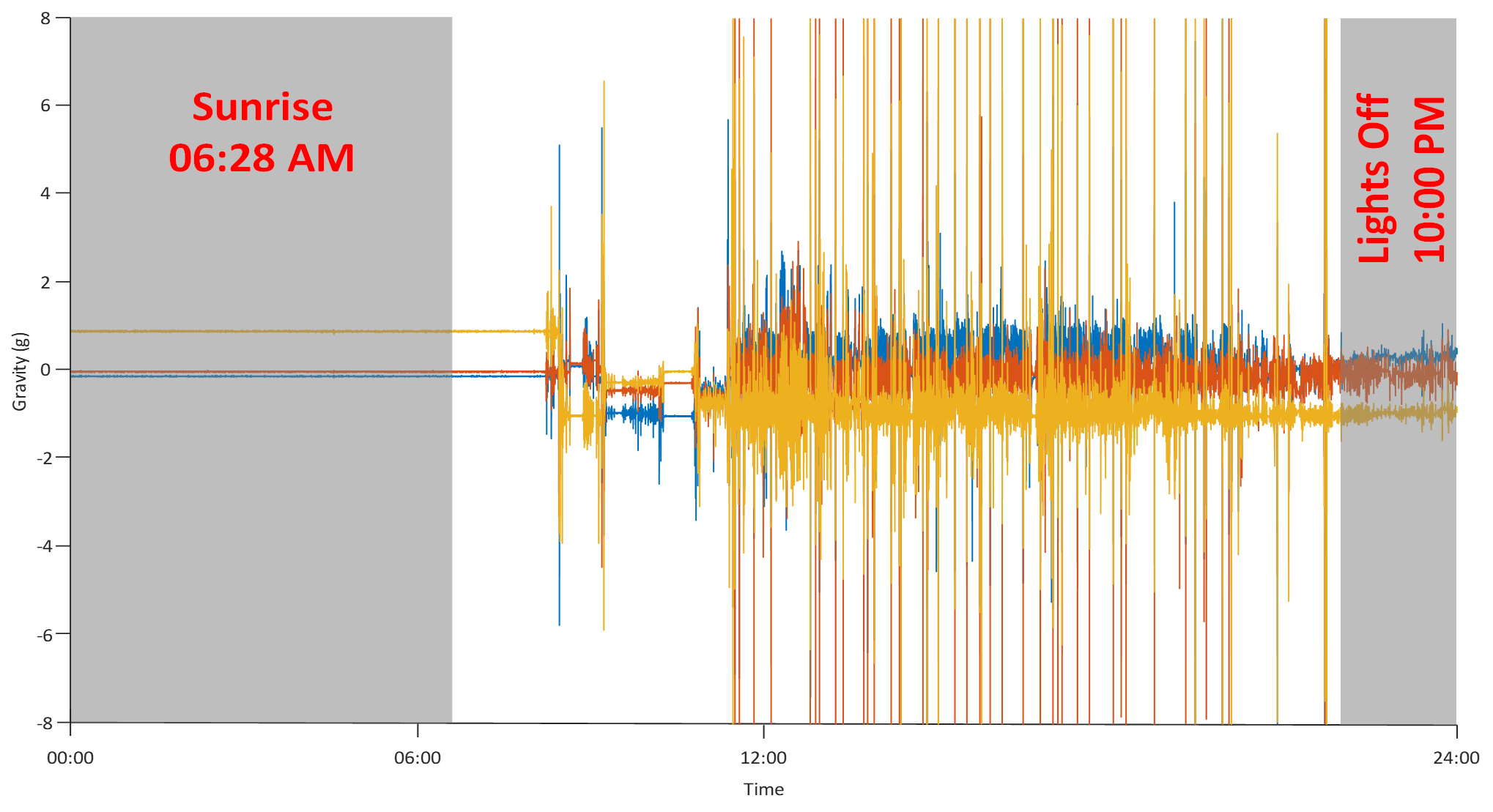}
\caption{Twenty-four hours time series chicken data.}~\label{fig:fig_10}
\end{figure}

In this section, we study the behavior of a chicken over the course of 24 hours and run our chicken behavior dictionary against the day-long dataset. The data corresponds to 11/22/2017 from midnight to midnight. The dataset is shown in Figure \ref{fig:fig_10}; and is of size 8,665,227 x 3 datapoints. The gray shaded regions correspond to midnight to sunrise (i.e. 06:28 AM) and the time artificial lights are turned off (i.e. 10:00 PM) to midnight.

\begin{figure}[h]
\centering
\includegraphics[width=\textwidth]{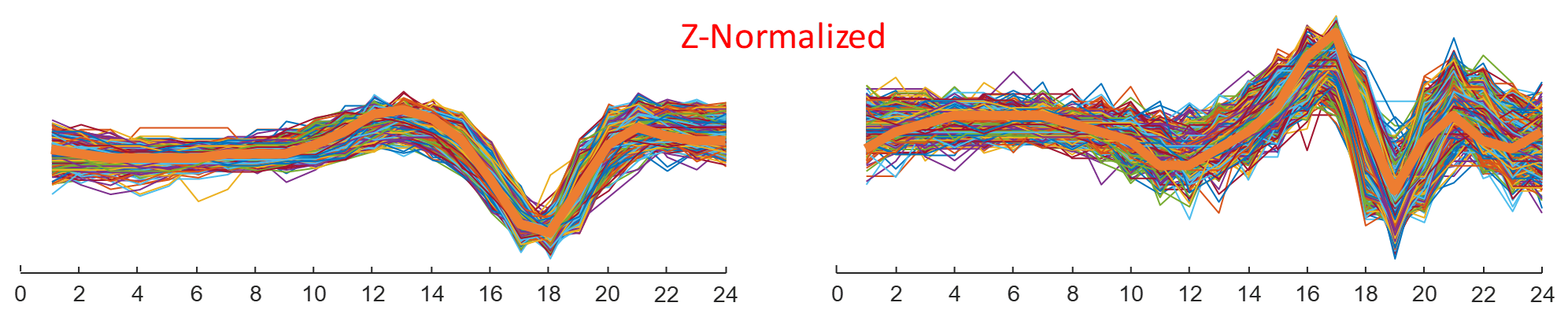}
\caption{The X-axis and Z-axis query-templates, along with matching Z-Normalized subsequences for feeding/pecking behavior in the twenty-four hours chicken dataset.}~\label{fig:fig_12_1}
\end{figure}

\begin{figure}[h]
\centering
\includegraphics[width=\textwidth]{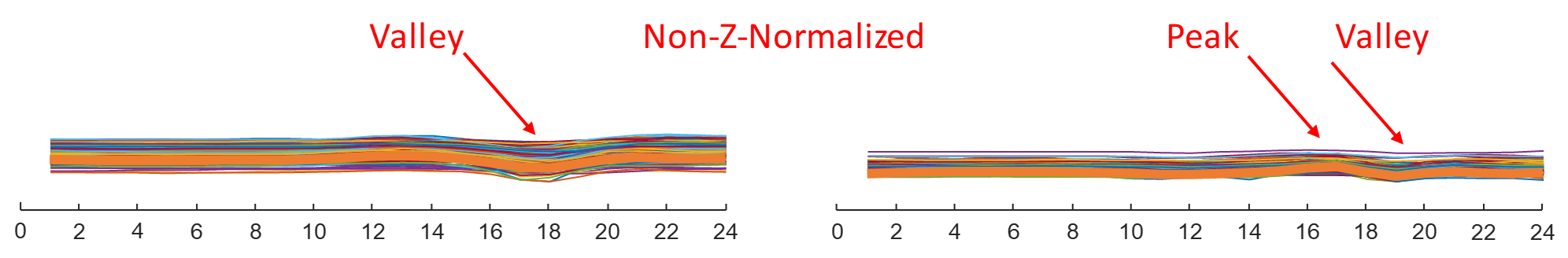}
\caption{The X-axis and Z-axis query-templates, along with matching Non-Z-Normalized subsequences for feeding/pecking behavior in the twenty-four hours chicken dataset.}~\label{fig:fig_12_2}
\end{figure}

\begin{figure}[h]
\centering
\includegraphics[width=0.8\textwidth]{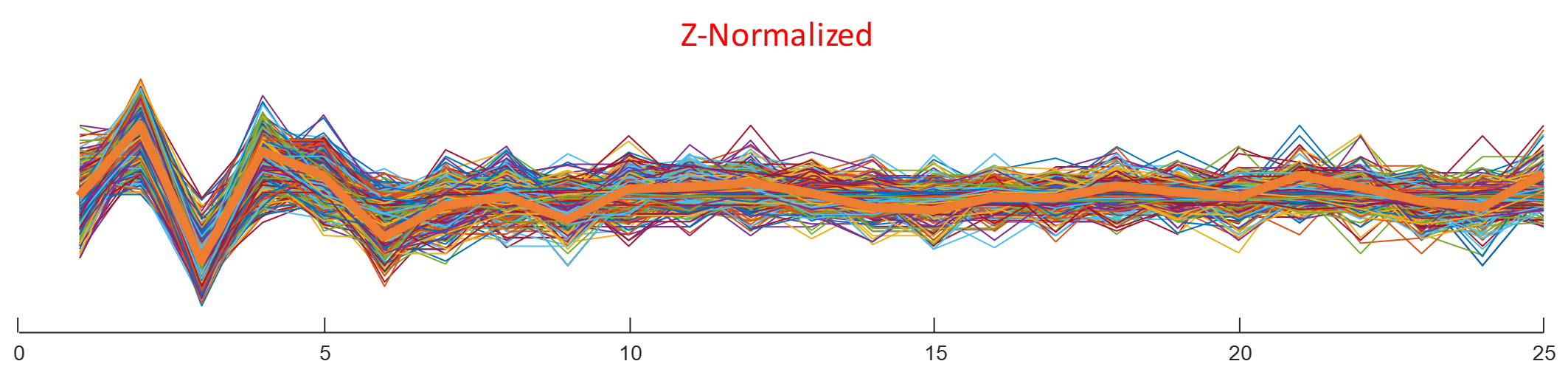}
\caption{The Z-axis query-template and matching Z-Normalized subsequences for the preening behavior in the twenty-four hours chicken dataset.}~\label{fig:fig_13_1}
\end{figure}

\begin{figure}[h]
\centering
\includegraphics[width=0.8\textwidth]{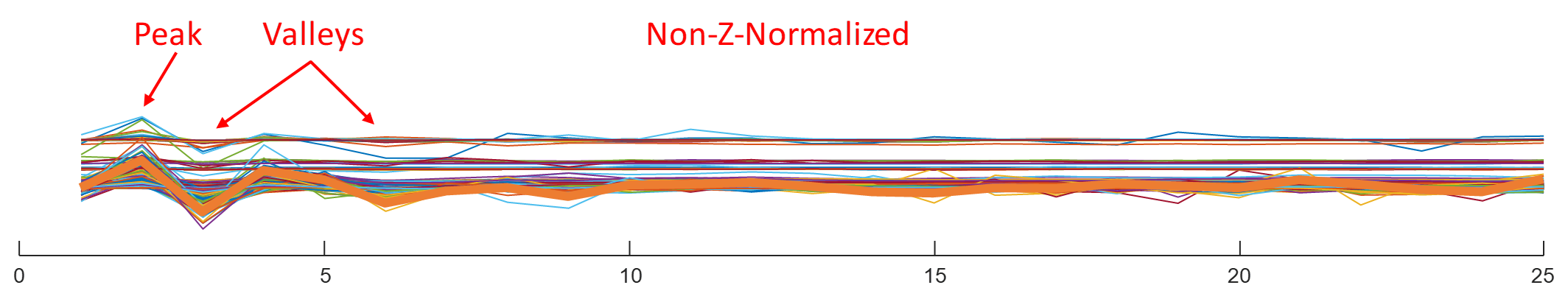}
\caption{The Z-axis query-template and matching Non-Z-Normalized subsequences for the preening behavior in the twenty-four hours chicken dataset.}~\label{fig:fig_13_2}
\end{figure}

\begin{figure}[!b]
\centering
\includegraphics[width=.8\textwidth]{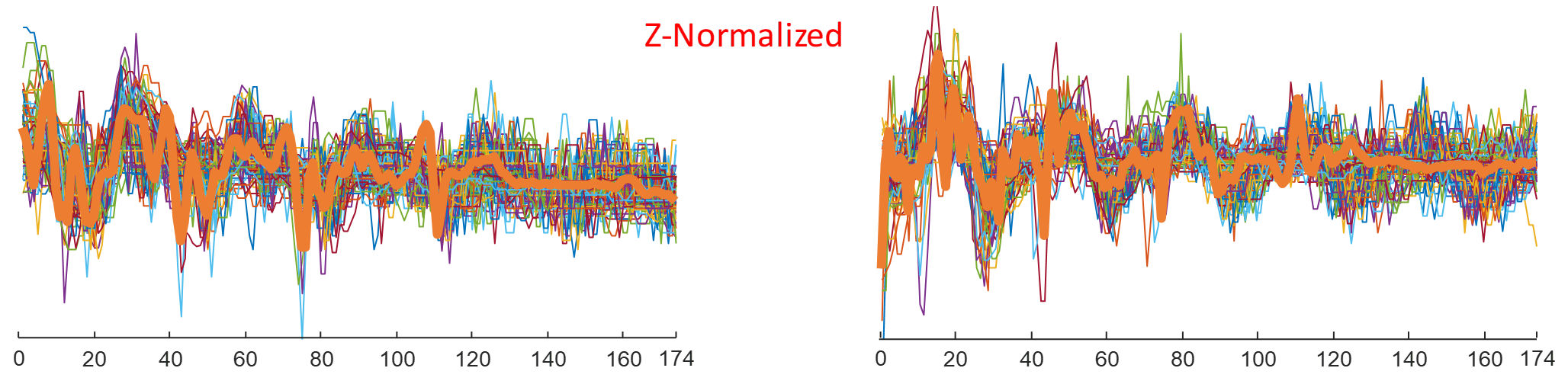}
\caption{The X-axis and Z-axis query-templates, along with Z-Normalized matching subsequences for dustbathing behavior in the twenty-four hours chicken dataset.}~\label{fig:fig_14_1}
\end{figure}

\begin{figure}[!t]
\centering
\includegraphics[width=.8\textwidth]{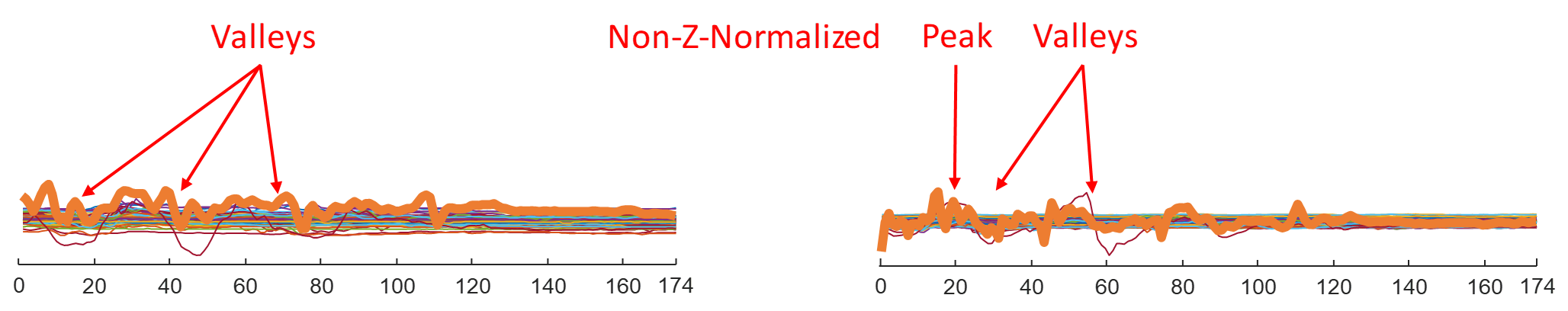}
\caption{The X-axis and Z-axis query-templates, along with Non-Z-Normalized matching subsequences for dustbathing behavior in the twenty-four hours chicken dataset.}~\label{fig:fig_14_2}
\end{figure}

We ran the dictionary of chicken behaviors against the twenty-four hours chicken dataset. Figures \ref{fig:fig_12_1} through \ref{fig:fig_14_2} present the matching subsequences for feeding/pecking, preening, and dustbathing query-template behaviors in the 24 hours chicken dataset.

Starting with Figures \ref{fig:fig_12_1} and \ref{fig:fig_12_2}, we show the matching subsequences in both the original and Z-normalized space to justify our choice of working with the Z-normalized representation \cite{mueen2017fastest}. Note that in the original space there are shifts in the mean that are inconsequential, yet which would dwarf the Euclidian distance calculations \cite{rakthanmanon2012searching}.

Looking at Figures \ref{fig:fig_13_1} and \ref{fig:fig_13_2} for the query-template and matching subsequences for the preening behavior, it can be seen that the matching subsequences have very high similarity with the query template at the beginning; however, we still see some dissimilarities along the rest of the query-template which may correspond to nibbling of the feathers with the beak. Finally in Figures \ref{fig:fig_14_1} and \ref{fig:fig_14_2} the matching subsequences for the dustbathing behavior are shown.

\begin{figure}[!b]
\centering
\includegraphics[width=0.9\textwidth]{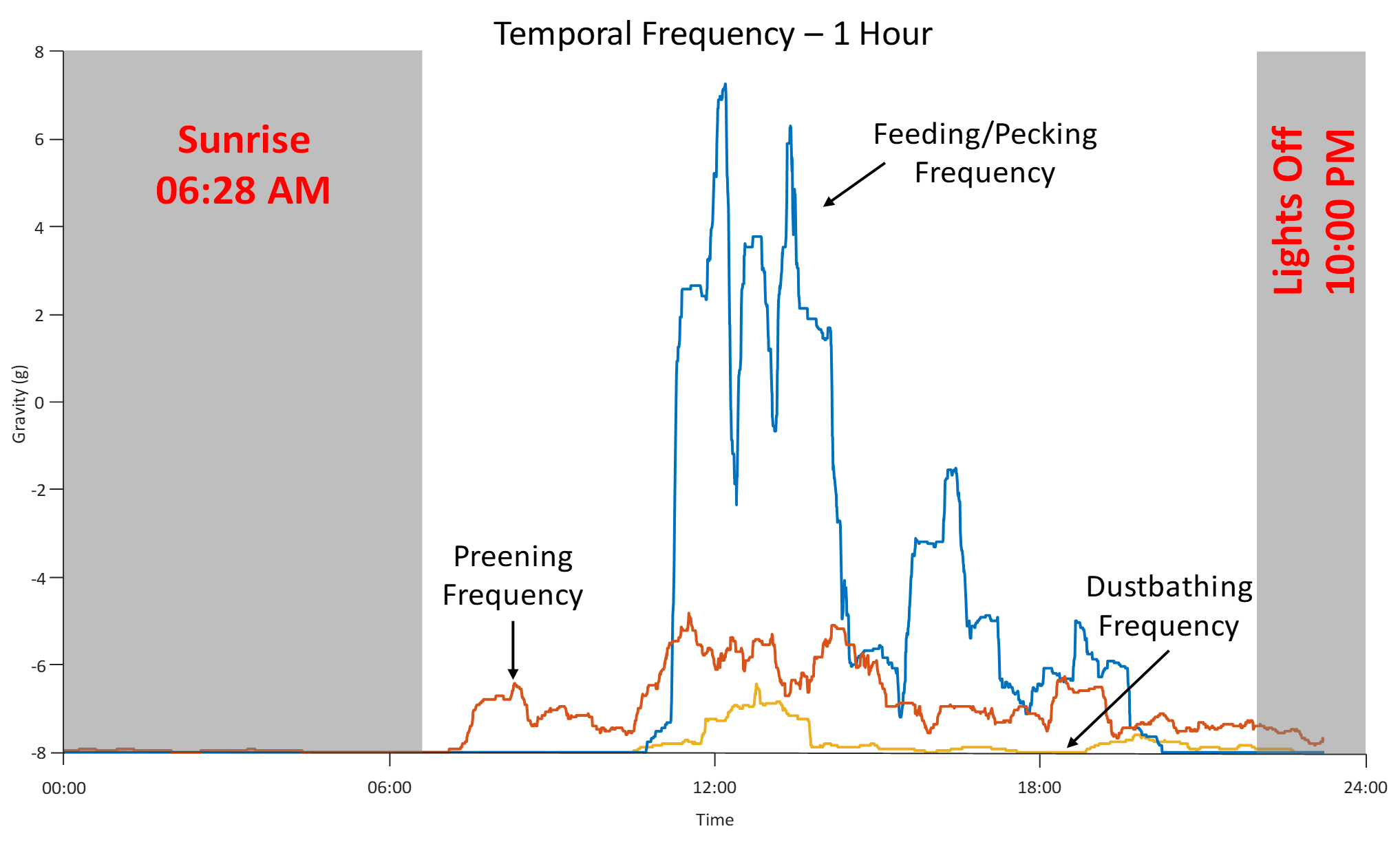}
\caption{Temporal frequency of chicken behaviors (feeding/pecking, preening and dustbating) for the twenty-four hour chicken dataset.}~\label{fig:fig_15}
\end{figure}

The matching subsequences are fairly well conserved, however, not as well as the feeding/pecking and preening behaviors. This can be due to the fact that dustbathing is a relatively long (appx. 1.74 s) and a rare behavior. Figure \ref{fig:fig_15} shows the frequency of each behavior over the entire 24 hours, as computed with a one-hour long sliding window.

As expected, the feeding/pecking behavior has the highest overall frequency; peaking between 11:00 AM and 15:00 PM. An interesting finding is that the chicken seems to begin the day with a preening session starting just after dawn. A more unexpected finding is that there is a “dip” in feeding that happens just after noon, and it seems to be replaced with an uptick in dustbathing.

\section{Conclusion}

In this chapter, we introduced an algorithm to learn a dictionary of behaviors from weakly labeled time series data. We demonstrated, with an extensive empirical study, that our algorithm can robustly learn from real, noisy and complex datasets, and that the learned query-templates generalize to previously unseen data. While our study was motivated by a pressing problem in poultry welfare, it could clearly be used to study cattle, goats, non-human primates \cite{mcfarland2013assessing}, or any other data source that presents itself as weakly labeled data. In future and ongoing work, we plan to examine much larger datasets; corresponding to several years of chicken data (recorded in parallel from multiple birds).

\chapter{Hybrid Shape \& Feature Classification: Fitbit for Chickens? Time Series Data Mining Can Increase the Productivity of Poultry Farms}

\newpage

\section{Introduction}

Poultry farming refers to raising domesticated birds (e.g. chickens, turkeys and ducks) to produce meat, eggs and/or feathers. Given the ever-growing world population and the demand for nutritious food, the poultry are farmed in great numbers. There are more chickens in the world than any other bird. Poultry meat and eggs contribute to human nutrition by providing high-quality protein and low levels of fat \cite{ciwf}.

In recent years, progress in sensor technologies have enabled practical deployments in increasingly prosaic settings. Recognition of human activities has been approached in various ways, among which the most practical and accurate are wearable sensors that are attached to the subject \cite{lara2012survey}. Such collected sensor data is typically extracted in the form of time series data; which can be investigated with a host of data mining techniques to summarize the behaviors of the subject \cite{abdoli2018time}. While there are hundreds of research efforts in time series classification, most studies on time series have only been evaluated on the highly contrived UCR Archive. In this work we show that real-world applications introduce unexpected difficulties and challenges that have been largely “hidden” by the curators of the UCR Archive (in fairness, the most recent version of the archive explicitly recognizes and addresses the issue). The two main issues we address are:

\begin{itemize}
    \item \textbf{The shape vs. feature classification dichotomy:} Most time series classification research efforts solely rely on either shape based or feature-based techniques \cite{dau2017matrix}\cite{moreau2009use}. However, as we show in this work there exist problems for which some classes are suited to just one of those paradigms, but some to the other. Thus motivated, we propose a joint shape/feature classification algorithm which outperforms both sole shape and feature based methods in terms of classification accuracy.

    \item \textbf{Learning from noisy and highly skewed real-world chicken data:} Most time series mining papers in the literature assume all classes are the same length, and the data is strongly labeled (again, simply reflecting what data is available in benchmarks).

\end{itemize}

\begin{figure}[h]
\centering
\includegraphics[width=.7\textwidth]{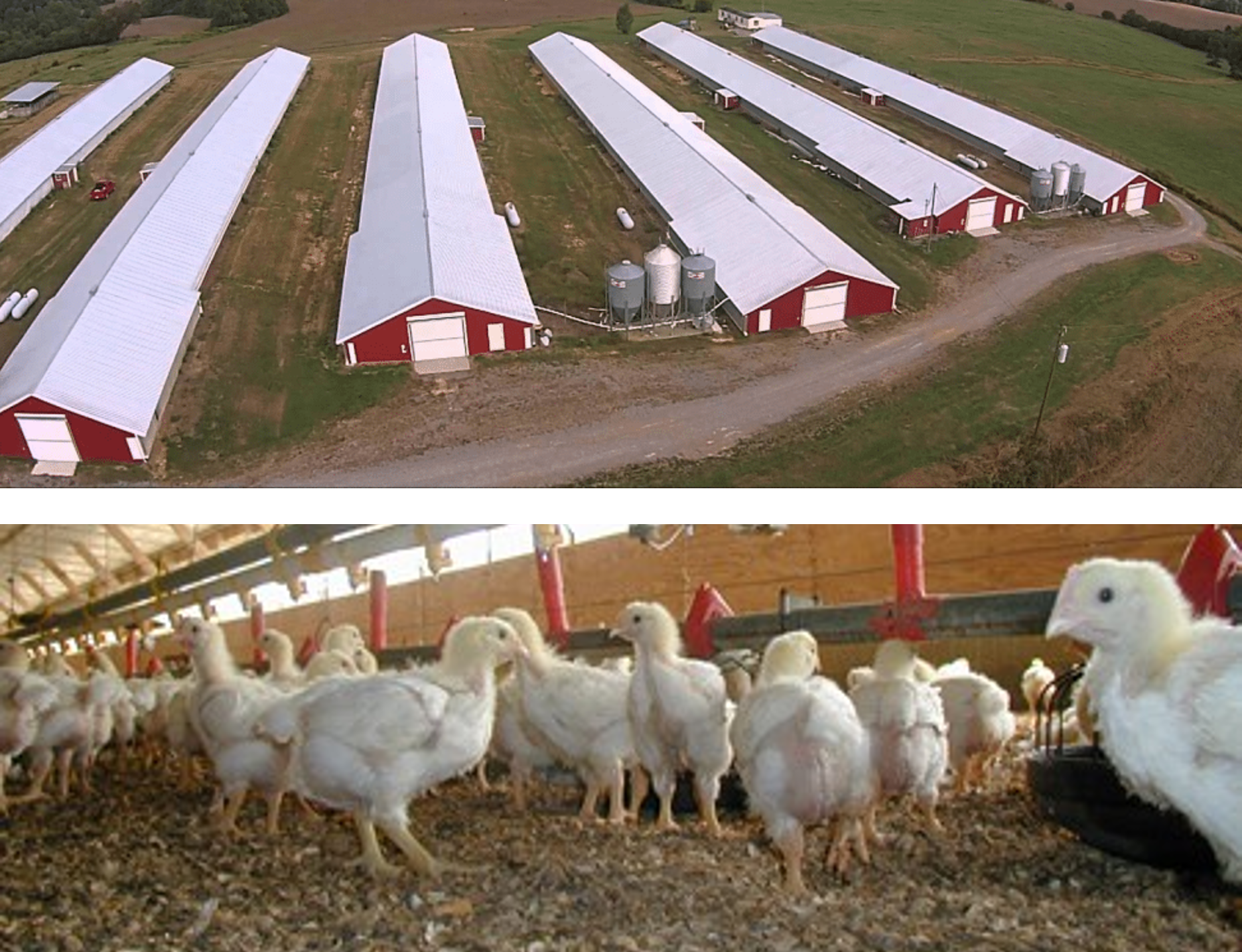}
\caption{(top) A poultry farm with six chicken houses in Cullman County, AL USA hints at the scale of poultry production (bottom) Chickens in a poultry house.}~\label{fig:kdd_2}
\end{figure}

In contrast, our domain of interest has well-defined behaviors that differ in length by almost an order of magnitude. Because of the two issues above, most algorithms in the literature are either not defined for our practical problem-at-hand, or they perform poorly.

\subsection{Poultry Management and Welfare}

The United States of America is the world’s largest poultry meat producer, with 18 percent of global output, followed by China, Brazil and the Russian Federation. Approximately 70 percent of chickens raised for meat globally are raised in intensive farming systems. This includes the majority of chickens in the US, UK, and Europe, as well as rapidly increasing numbers in developing countries \cite{FAO}, as shown in Figure \ref{fig:kdd_2}.

If poultry are to achieve their genetic potential for meat or egg production, they need an environment that meets their physiological requirements. This includes: (1) a suitable physical environment in terms of temperature, humidity, air movement and the surfaces on which they live; (2) adequate food and water; (3) minimal exposure to disease causing organisms; and (4) avoidance of exposure to stress resulting from the physical and social environment. The factors influencing these are determined largely by housing and management. Consumers increasingly want to be sure that all animals being raised for food are treated with respect and are properly cared for during their lives, and companies involved in raising chickens for food are becoming increasingly responsive to the public’s concern. They recognize that they have an ethical obligation to make sure that the animals on their farms are well cared for. The chicken industry has come together on a specific set of expectations that will ensure that the birds they raise are taken care of with the highest standards starting at hatch \cite{FAO_Housing}.

\subsection{Challenges in Working with Chickens}

There are hundreds of studies on quantifying human body behaviors with sensors \cite{reiss2011exploring}. Such studies typically involve finding discrete well-defined classes of behaviors, and then monitoring data for future occurrences of behaviors. One example is “step-counting” to measure compliance with a suggested exercise routine. However, the task of studying behaviors (activity recognition) in chickens is considerably more difficult than humans for the following reasons, as outlined in Table \ref{tbl:kdd_1} and discussed below:\\

\begin{table}[h]
\centering
\begin{tabular}{|l|l|l|}
\hline
Activity               & Humans       & Chickens        \\ \hline
Sensor Placement       & Flexible     & Limited         \\
Individual Variability & Well-Studied & Limited-Studies \\
Cooperativity          & High         & Limited to None \\ \hline
\end{tabular}
\caption{Activity recognition trends in humans versus chickens}~\label{tbl:kdd_1}
\end{table}

\vspace{-0.25in}

\begin{itemize}
    \item In case of humans, the sensors can be easily placed on the extremities of the limbs (i.e. smart-shoes or smartwatches); However, the placement of sensors on chickens has been primarily restricted to the back of the birds due to sensor limitations and the welfare of animal. This provides only coarse information about the bird’s behaviors.

    \item The variability of human behaviors is well-studied, and it is understood what fraction can be attributed to individual’s personality, sex, body type, mood and so forth, versus variability in sensor placement \cite{doppler2009variability}. More importantly, it is known how to account for this variability \cite{kreil2016coping}. However, it is less clear how much variability exists in birds and how we can best account for it \cite{abdoli2018time}.

    \item Creating a dictionary for a human subject is relatively straight-forward. During an explicit training session, behaviors of interest can be “acted out” in a fixed order for a fixed duration of time. For example, one of the most studied human motion time series datasets is the UCR archives gun-point \cite{ding2008querying}. When recording that dataset, the actor’s behaviors were cued by a metronome \cite{dau2017matrix}, to enable subsequent extraction of the behaviors. Chickens are clearly not as cooperative, and many hours/days of video recordings must be analyzed to create a behavior dictionary. Moreover, it is difficult, even for an experienced avian ethologist, to define precisely where a given behavior begins and ends, thus we must be able to work with “weakly labeled” data. For example, our data may be labeled as “most of that eight seconds is comprised of dustbathing” or “there are about twelve pecks in that ten second snippet”. 
\end{itemize}

In this work \cite{abdoli2020fitbit}, we introduce a novel shape/feature classification algorithm, which can take such weakly labeled data together with some mild constraints “a preening behavior probably lasts between 0.3 and 1.5 seconds” and automatically learns to classify the behaviors.

It is important to note that we are not advocating placing sensors on every chicken in the flock. This would clearly be prohibitively expensive and time consuming. Our claim is that by studying the behaviors of a few sentinel birds, we can utilize the learned lessons to improve the welfare for the entire breed. This idea is widely understood, so we will not dwell on it further. See \cite{dawkins2004chicken} and the references thereof.

\section{Related Work}

As noted above, the many thousands of research efforts on classification of human behavior do not bear directly on the task-at-hand, thus we confine our discussion to non-human animals. In recent works \cite{walton2018evaluation}\cite{barwick2018categorising}, sensors were used for classification of sheep behaviors; with mounted sensors on ear/collar or leg of the sheep. Similar work has been performed for domestic bovines \cite{smythe2015behavioral} and goats \cite{moreau2009use}, and for various kinds of wild animals. There has been a work for monitoring of insects \cite{batista2011sigkdd}. In addition, there has been some work on poultry behaviors using sensors. However, this work is complementary to our efforts. They use only statistical features (mean, entropy, etc.) to quantify periods of general behaviors, such as sleep, stand, walk, etc. \cite{banerjee2012remote}. In contrast, Abdoli et al. \cite{abdoli2018time} used the shape of the time series to annotate very specific and dynamic behaviors, such as individual instances of a single peck. However, as we will show in a later section, relying solely on shape will result in low accuracy, as some chicken behaviors (for example, preening) have highly conserved values for some features, but are not expressed in well conserved shapes.

\section{Data Acquisition and Cleansing}

\begin{figure}[h]
\centering
\includegraphics[width=.5\textwidth]{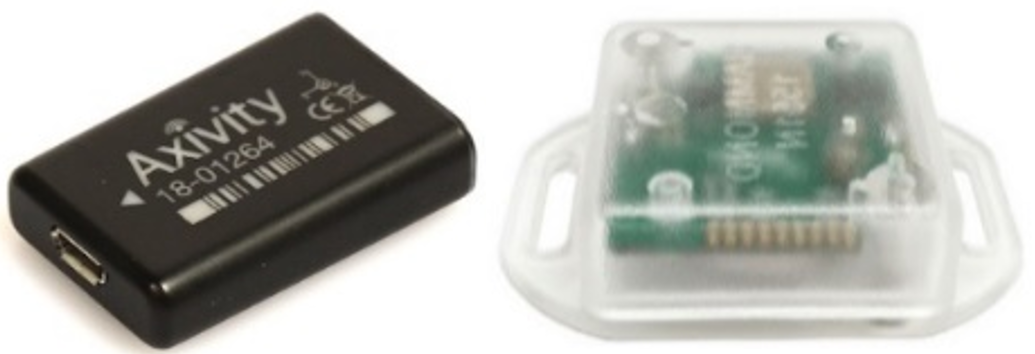}
\caption{(left) Axivity AX3 data logger sensor used for batch processing mode (right) MbientLab Meta Tracker (MTR) sensor used for online real-time processing mode.}~\label{fig:kdd_3}
\end{figure}

All chickens were housed and cared for in accordance with UC Riverside Institutional Animal Care and Use Protocol. Data is collected from chickens by placing the sensor on bird’s back. Placement of the sensor in this allows for high-quality recording chicken behaviors, while the minimum interference and discomfort. We initially collected data using the Axivity AX3 data logger sensor \cite{abdoli2019time}, shown in Figure \ref{fig:kdd_3} (left), weighing about 11 grams and configured with 100 Hz sampling frequency and +/- 8g sensitivity which allows for two weeks of continuous data collection with the battery fully charged.

\begin{figure}[t]
\centering
\includegraphics[width=.8\textwidth]{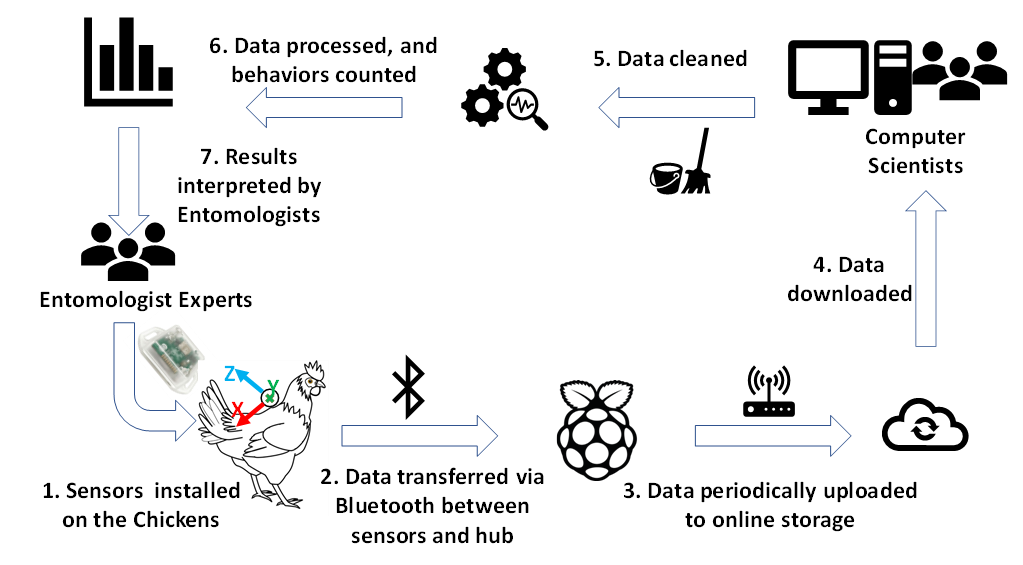}
\caption{The workflow to study chicken behavior and welfare in online processing mode.}~\label{fig:kdd_4}
\end{figure}

As a data logger, the Axivity AX3 saves all acquired data to an internal memory device which can only be accessed after the sensors are removed from the chickens. This requires considerable human time and effort. On the other hand, diseases and abnormalities might spread quickly throughout the chickens in the houses. Therefore, in order to provide instantaneous access to the chicken data we utilized MbientLab MetaTracker sensors, shown in Figure \ref{fig:kdd_3} (right), which are capable of wireless real-time data transmission through the Bluetooth technology. In Figure \ref{fig:kdd_4} we show the logic model for our operations. The workflow, as outlined goes from step 1 to 7.

Note that the workflow was supposed to be linear, steps 1 to 7. However, as with most real-word deployments, as we approached step 7, the computer scientists in our group saw opportunities to improve the quality of the data, and the entomologists saw opportunities to monitor additional behaviors. Thus, we cycled through this workflow several times on new cohorts of birds  before converging on a commercially deployable system. This also might be a good place to note why entomologists, and not directly bird experts, are the main customers of our system.

By far, the most important stressor of chickens, and most important vectors of chicken diseases are insects and mites \cite{dawkins2004chicken}\cite{brown1974effect}.

\section{Definitions and Notations}

We begin by providing definitions and notation to be used throughout the paper.

\textbf{Definition 1:} A time series $T$ is a sequence of real-valued numbers $t_i$: T = $[t_1, t_2, …, t_n]$ where $n$ is the length of $T$.

We are typically not interested in the global properties of time series, but in the similarity between local subsequences:

\textbf{Definition 2:} A subsequence $T_{i,m}$ of a time series $T$ is a continuous subset of the values from $T$ of length m starting at position $i$. $T_{i,m}$ = $[t_i, t_{i+1}, …, t_{i+m-1}]$ where $1 \leq i \leq n – m + 1$.

We will classify the time series using a combination of feature and shape measures.

\textbf{Definition 3:} The nearest neighbor (NN) classifier for a time series $T$ is an algorithm that for each query $Q$ (i.e. subsequence) finds its nearest neighbor in $T$ and assigns $Q$ that neighbor’s class.

In case of a shape-based classifier, the nearest neighbor is defined using either the Euclidean distance (ED) or the Dynamic Time Warping (DTW) as the distance measure. The distance between a query (i.e. an unlabeled exemplar) and all the other subsequences in the time series is stored in an ordered vector called the distance profile ($D$).

\textbf{Definition 4:} A distance profile $D$ is a vector of Euclidean distances between a given query and each subsequence in the time series.

To avoid extracting redundant subsequence matches in distance profile we must be aware of trivial matches:

\textbf{Definition 5:}  Given a time series $T$, containing a subsequence $T_{p, m}$, if $T_{p,m}$ scores highly on any scoring function, then $T_{j, m}$ will almost certainly score high on the same function. These spurious high scoring subsequences are trivial matches.

To avoid counting trivial matches when finding matches to a query, we discard some of the patterns using the concept of an exclusion zone, a standard practice \cite{chiu2003probabilistic}.

While distance is the measure of similarity between two subsequences in a shape-based classifier, the feature-based classifier finds similarity based on a set of features.

\textbf{Definition 6:} Given time series $T$ and  function $X$, the feature vector $F$ will be the value of $X$ for every subsequence of $T$. Each feature vector $F$ corresponds to a measurable property of the time series.

\section{Methodology}
The problem of time series classification has been around for decades. There are two main approaches for time series classification in the literature, namely, shape-based classification \cite{gharghabi2018matrix}\cite{imani2018matrix}\cite{imani2019matrix} and feature-based classification \cite{lubba2019catch22}. Shape-based classification determines the best class according to a distance measure (e.g. Euclidean distance). Feature-based classification, on the other hand, finds the best class according to the set of features defined for the time series. These features measure properties of the time series (e.g. autocorrelation, complexity, etc.) \cite{alaee2020features}.

\subsection{Shape versus Feature Classification}

As noted above, researchers typically use either shape-based or feature-based techniques for classification of time series. However, deciding which classifier is the best for the problem at hand is a difficult problem to solve. There are three obvious possibilities.

\begin{itemize}
    \item One of the two techniques dominates for all problems. However, an inspection of the literature, or a little introspection convinces us otherwise. There are clearly problems for which one of the two approaches is much better on all classes \cite{alaee2020features}.

    \item On a problem-by-problem basis, one of the techniques dominates. For example, perhaps for electrocardiograms shape is the best, but for electroencephalograms feature is the best. This seems to be the implicit assumption of most of the community \cite{fulcher2013highly}.

    \item On a single problem, it might be possible that one of the techniques is better for one subset of the classes, and the other technique is better for another subset.
\end{itemize}

Given this third possibility, it is clear that neither of the two techniques will dominate for some problems, but that some “combination” of both might be the best.

We have observed that the task of classifying chicken behavior from accelerometer data is unlikely to yield to a single modality of classification. The pecking behavior has highly conserved shape. Whereas the preening and dustbathing behaviors do not have a stereotypical shape, but are recognizable using several features, including “complexity” and “frequency” \cite{fulcher2013highly}.

\subsection{Chicken Behavior Classification}

From literature reviews \cite{li2020automated}\cite{brown1974effect}, and conversations with poultry experts, we expect that the following behaviors correlate with poultry health:

\begin{itemize}
    \item \textbf{Feeding/pecking:} bringing the beak to the ground and retrieving a morsel of food.
    \item \textbf{Preening:} grooming of the feathers using the beak.
    \item \textbf{Dustbathing:} sitting or rolling in the dirt.
\end{itemize}

In particular, birds infected with ectoparasites are expected to do more preening/dustbathing \cite{dawkins2004chicken}. Table \ref{tbl:kdd_2} details the combination of shape and features we used for classifying pecking, preening and dustbathing behaviors. 
In case of the pecking behavior the shape works well, however, to distinguish and ward off noisy subsequences that look like pecks we further applied standard deviation and complexity features. The dustbathing behavior (likewise preening) does not have a well-conserved shape. So, we utilized features to classify their instances. Given the intense nature of the preening and dustbathing behaviors the “complexity” feature is useful to distinguish the aforesaid behaviors from other behaviors \cite{batista2014cid}. Moreover, to further classify between preening and dustbathing instances we applied the power spectral density feature which provides discrimination to distinguish these behaviors \cite{fulcher2013highly}. Note that although we used the complexity feature for the pecking, preening and dustbathing, we learn different thresholds for each behavior class.

As can be seen in Table \ref{tbl:kdd_2}, the length for pecking behavior instances is assumed to be constant, while the length for the instances of preening and dustbathing instances can vary from 0.3 of a second to 8 seconds. The shape-based classification will not work well for preening and dustbathing and we should utilize features (i.e. complexity and power spectral density).

\begin{table}[h]
\centering
\begin{tabular}{|l|l|l|l|}
\hline
Behavior        & Length       & Shape                                      & Feature        \\ \hline
Pecking         & Constant      & \checkmark (ED)            & \checkmark (SD, CMP) \\
Preening        & Variable      & X                                          & \checkmark (SD, SPD) \\
Dustbathing     & Variable         & X                                       & \checkmark (SD, SPD) \\ \hline
\end{tabular}
\caption{Shape-Feature Classification for Chicken Behaviors. Keys: ED = Euclidean Distance, SD = Standard Deviation, CMP = Complexity, SPD = Spectral Power Density.}~\label{tbl:kdd_2}
\end{table}

We are now in a position to discuss the proposed combined shape-feature algorithm.

\subsection{Proposed Approach}

\begin{enumerate}
    \item \textbf{Propose a set of useful features and shapes}, allowing the possibility that different classes may best be distinguished with different subsets of shapes and features.
    \item \textbf{Calculate the shape-vectors and feature-vectors}, given the identified set.
    \item \textbf{Learn relevant thresholds for every class} (given the user’s class-dependent tolerance for false positives/false negatives).
\end{enumerate}

The algorithms outlined above have two subroutines outlined in Table \ref{KDD_ALG_1}. Individual elements are motivated and explained in the following subsections.

\textbf{Propose a set of useful features and shapes:} The user will provide a set of shapes and features for every class. These suggested shapes and features can come from domain experts or visual inspection of the time series by the user. The choice of which features to use is beyond the scope of this paper. Fulcher et al. \cite{fulcher2013highly} presented a library of over 9,000 features that can be used to quantify time series properties including the classic features such as min, max, standard deviation, periodicity etc.

\textbf{Calculate shape and feature-vectors:} We are given a time series, the set of classes describing the time series and the set of shapes and features describing each individual class. The classification process involves calculating shape-vector and feature-vector(s), based on the proposed set of shapes and features, for every class.

\textbf{Set relevant thresholds for every class:} For each class, we need a threshold that best defines our relative tolerance for the false positives vs. false negatives. The thresholds can be either manually adjusted by the user (static) or learned through a feedback loop (dynamic), assuming ground truth labels are available. In the dynamic case, the user may inspect the results produced by multiple runs of the algorithm and choose the threshold setting corresponding to the most desired point on the ROC curve.

Algorithms \ref{KDD_ALG_1} and \ref{KDD_ALG_2} show calculation of shape and feature vectors for pecking, preening and dustbathing behaviors. The typical shape of an instance of pecking behavior is shown in Figure 8. Note that, in principle, a single behavior could have two or more possible instantiations; just like the number four has two written versions, closed ‘4’ and open ‘4’, which are semantically identical. We call such behaviors a polymorphic behavior. Our algorithm allows for having multiple instances of shapes for the same class, so we can account for instances which belong to the same class but differ in shape.


\begin{algorithm}
\textbf{Algorithm for computing shape-based profile and feature-vectors for pecking behavior}
\label{KDD_ALG_1}
\begin{figure}[h]
\centering
\includegraphics[width=\textwidth]{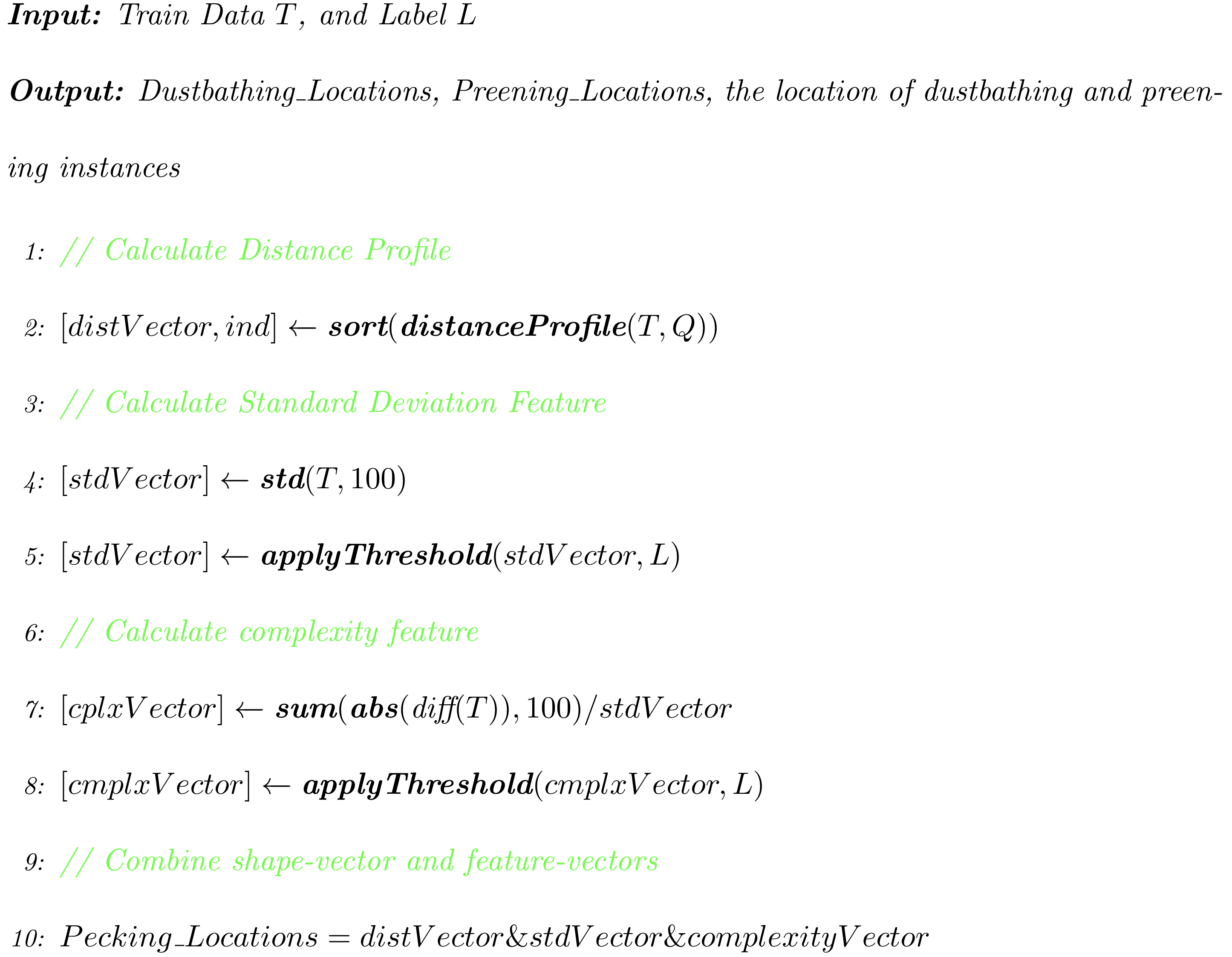}
\end{figure}
\end{algorithm}

In line 2, the distance profile (Definition 4) of the shape-based query for pecking behavior is calculated and sorted in ascending order. Essentially, the subsequences most similar to the query Q are prioritized as likely candidates. In \cite{abdoli2018time} the authors attempted to classify just with shape, however, given technical limitations of time series data (noisy data and etc.) and fast-paced nature of pecking behavior (lasting for one fourth of a second) the authors had to set a conservative threshold to avoid misclassifications of real chicken pecks and noisy data that look similar to pecks. In this study, we mitigate this issue with combined shape-feature classification. Standard deviation and complexity are the two suggested features to distinguish a real peck from similar noisy data.

In line 4, we calculate the standard deviation over the time series T with a sliding window of one second (data was recorded at 100Hz). Next, in line 5, stdVector is passed to applyThreshold function which tests different thresholds given the label L and finds the best threshold for the feature vector. Similarly, in line 7, the complexity feature is calculated over the time series with a sliding window of one second. cmplxVector is passed to the applyThreshold function to obtain the threshold yielding the best classification results for the feature complexity. Given the distance profile (distVector), standard deviation vector (stdVector) and complexity vector (cmplxVector), in line 10, we combine the shape-feature vectors using logical operations so that noisy data are filtered by standard deviation and complexity features.

Table \ref{KDD_ALG_2} shows the steps for classification of preening and dustbathing behaviors using the complexity and power spectral density features. In line 2 the complexity feature is calculated over the time series. In line 3 cmplxVector and class labels L are passed to applyThreshold so that the best threshold for the preening and dustbathing classes are found.

\newpage


\begin{algorithm}
\textbf{Algorithm for computing feature-vectors for preening and dustbathing behaviors}
\label{KDD_ALG_2}
\begin{figure}[h]
\centering
\includegraphics[width=\textwidth]{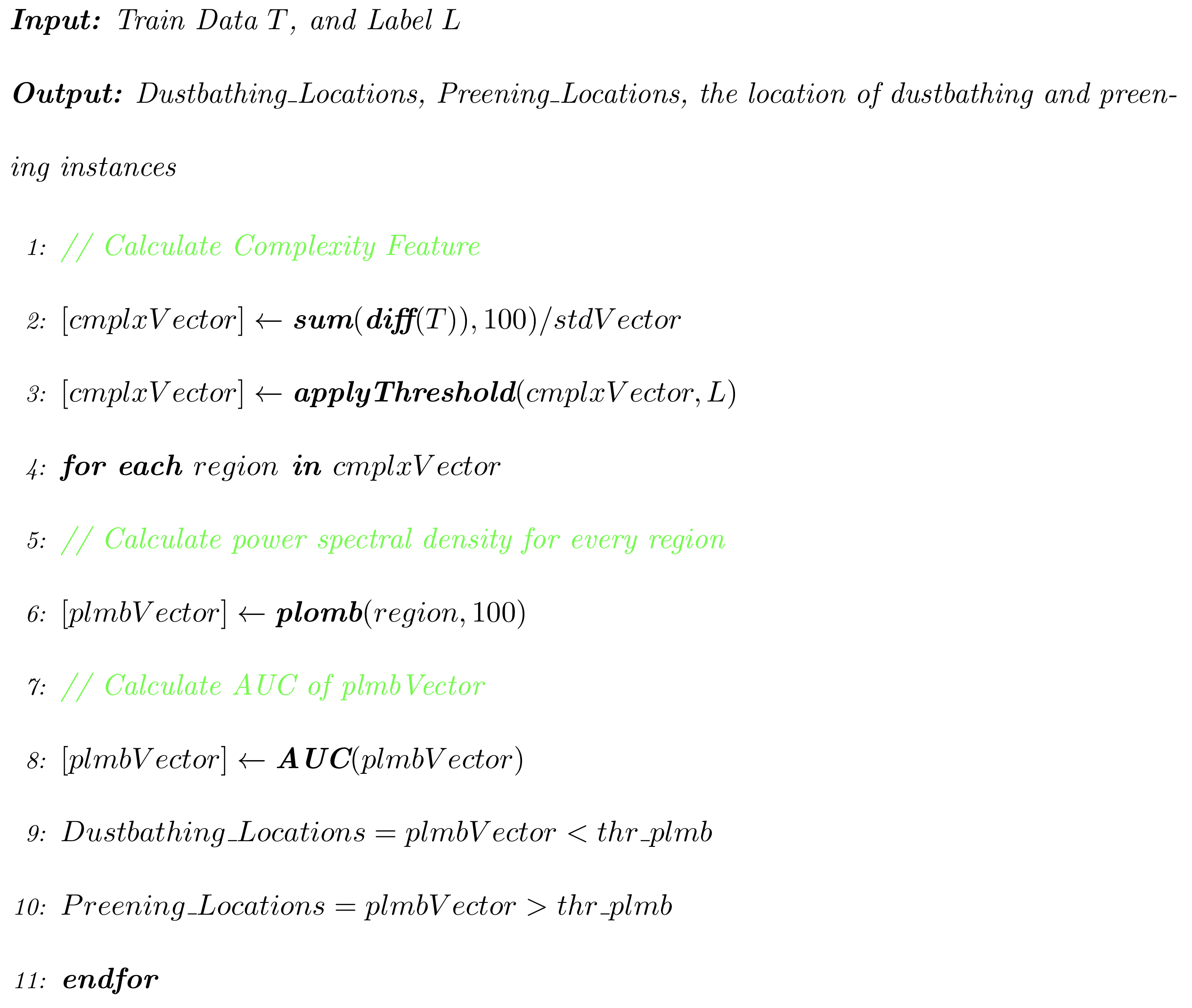}
\end{figure}
\end{algorithm}

\begin{figure}[h]
\centering
\includegraphics[width=\textwidth]{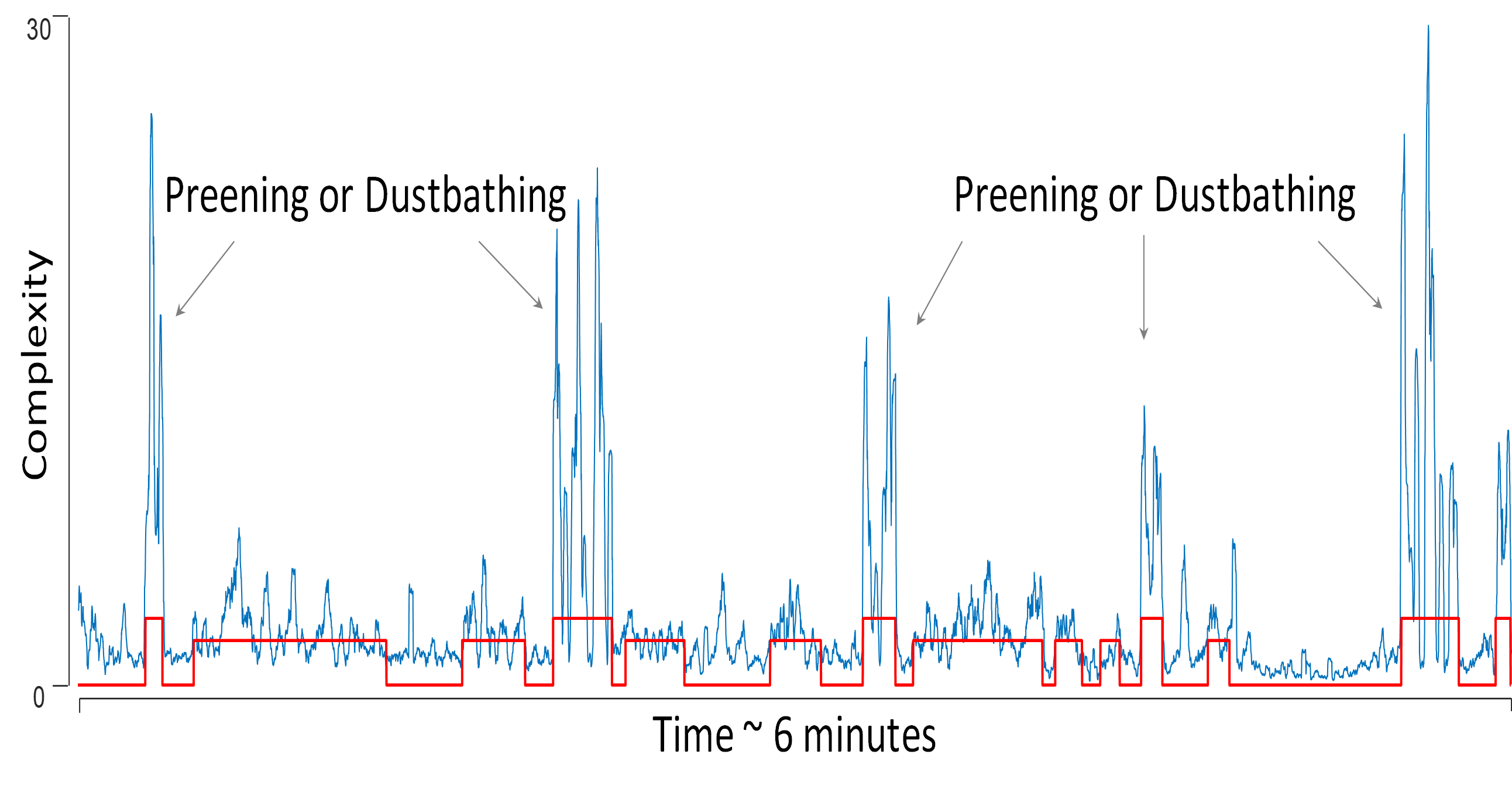}
\caption{The complexity feature vector for distinguishing preening and dustbathing instances from other classes. Note that the feature (\textcolor{blue}{blue}) tends to peak at location marked as positive with the vector of class labels (\textcolor{red}{red}).}~\label{fig:kdd_5}
\end{figure}

Figure \ref{fig:kdd_5} depicts the complexity feature vector which as demonstrated in the figure helps with distinguishing preening and dustbathing behaviors from other classes. Returning to Algorithm \ref{KDD_ALG_2}, following the calculation of complexity feature vector, we identify the regions corresponding to either preening or dustbathing behavior.

\begin{figure}[h]
\centering
\includegraphics[width=\textwidth]{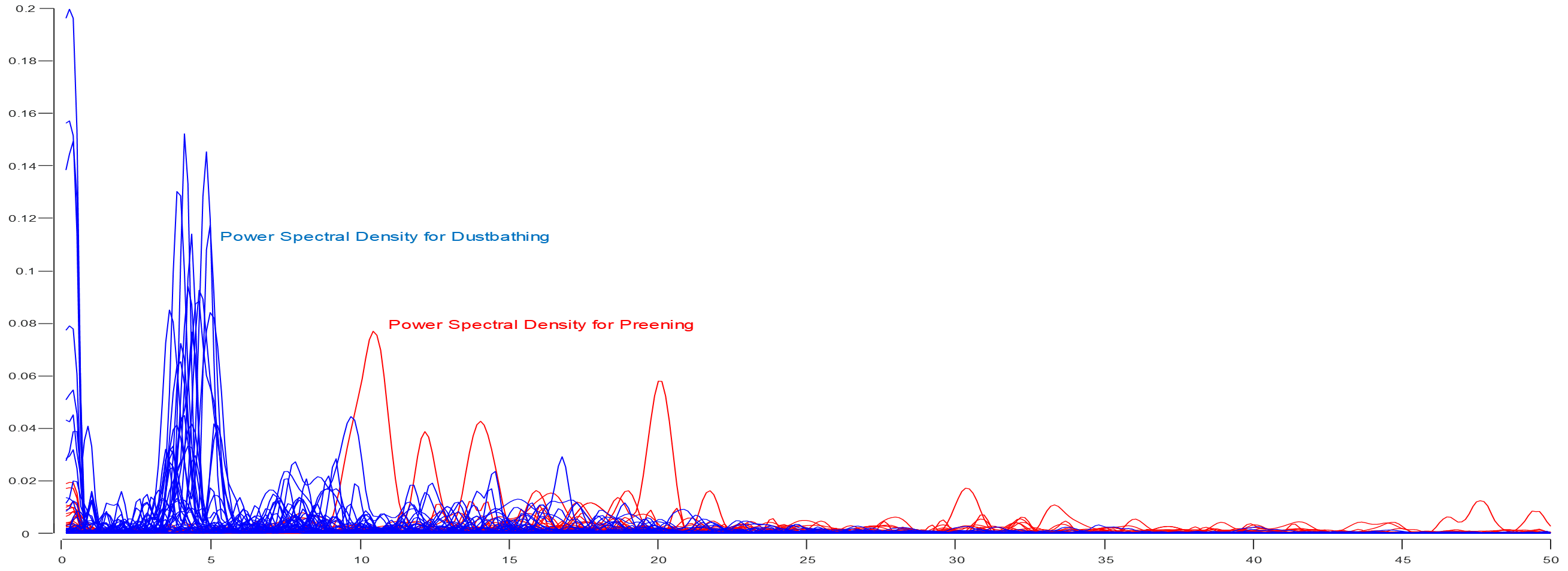}
\caption{The power spectral density feature vector for distinguishing between preening and dustbathing instances.}~\label{fig:kdd_6}
\end{figure}

In line 4, we initiate a for loop and for every distinguished region from the complexity feature we calculate the power spectral density function over that region, as shown in Figure \ref{fig:kdd_6}. The main benefit of the spectral power density is to differentiate between preening and dustbathing instances. Next, we account for the area under the curve (AUC) of the power spectral density of every instance, and the value of the AUC helps to classify between preening and dustbathing instances.

Following discussion of the shape-feature based classification algorithm we will initially provide extensive evaluation results for a labeled chicken dataset. Next, we will demonstrate that the proposed classification algorithm generalizes well to unforeseen chickens for which we do not have any labels of any kind.

\section{Experimental Evaluation}

To ensure that our experiments are reproducible, we have built a supporting website; which contains all data, code and raw spreadsheets for the results. Initially we present the results for the training/test dataset, shown in Figure \ref{fig:kdd_7}, and then we proceed to a case study of utilizing the shape-feature based classification algorithm to recognize healthy and unhealthy chickens based on the behavior count throughout the day.

\subsection{Performance Evaluation}

\begin{figure}[h]
\centering
\includegraphics[width=\textwidth]{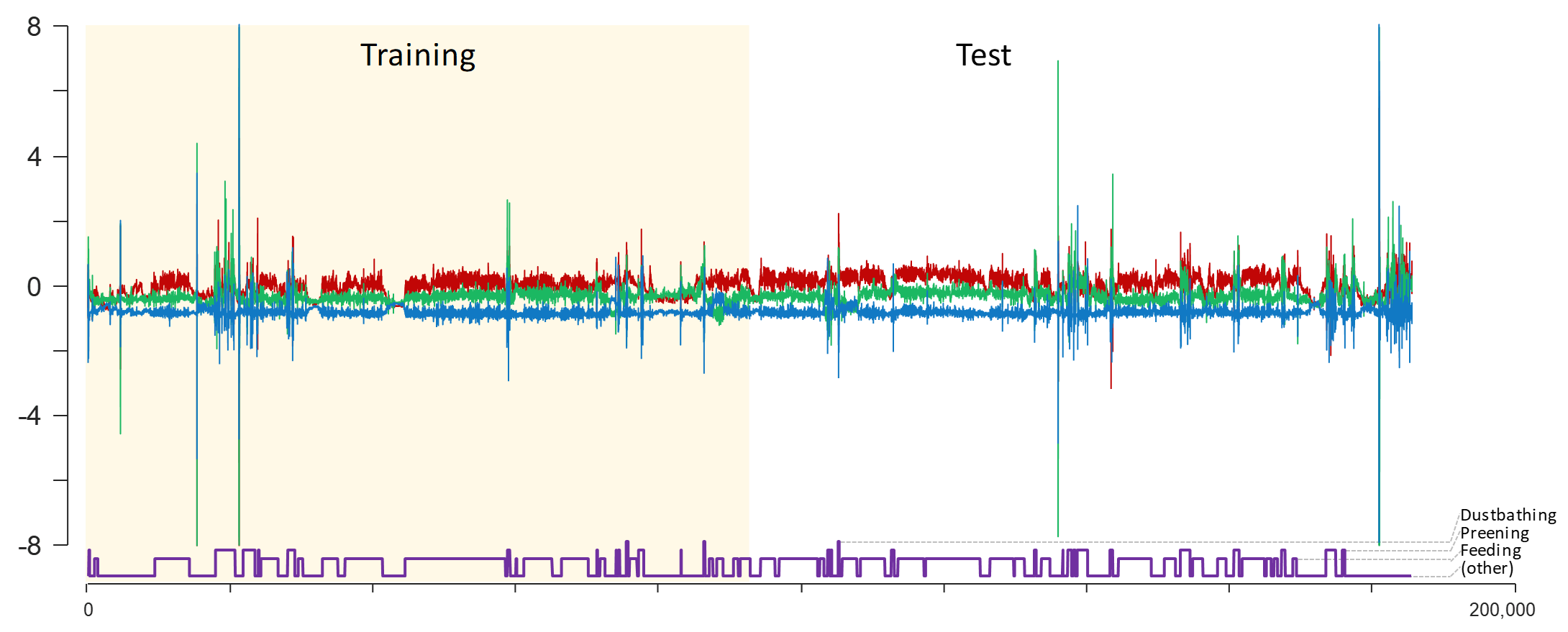}
\caption{Three-dimensional chicken time series (the top/red time series is X-axis; middle/green time series is Y-axis and bottom/blue time series is Z-axis time series). The purple lines represent annotations of observed chicken behaviors captured on video; the height of each annotated region represents a distinct behavior.}~\label{fig:kdd_7}
\end{figure}

The original dataset is split into mutually exclusive training and test datasets, as illustrated in Figure \ref{fig:kdd_7}. For some fraction of the time series data collected, a video camera also recorded the chicken activity (approximately 30 minutes).

This video recording provides ground truth to act as training data. The sensor data was carefully annotated \cite{ELAN}, based on the video-recorded chicken activities; it must be noted that even the most careful human labeling of chicken behaviors can contain errors, especially false negatives.

As shown in Figure \ref{fig:kdd_7}, the annotations (purple) take the form of a categorical vector that indicate that in the corresponding region one or more examples of the corresponding behavior were observed. Such data is often called “weakly labeled” data. In addition, there are almost certainly instances of the behavior outside the annotated regions which the annotator failed to label, perhaps because the chicken in question was occluded in the video. However, we believe that such false negatives are rare enough to be ignored, and moreover, our algorithm is not very sensitive to mislabeled data.

We do not know the exact number of instances of a class inside a region. As noted above, the labels are of the form “there are about ten pecks in that six seconds snippet”. To address this issue, we utilize the concept of Multiple Instance Learning (MIL) \cite{amores2013multiple}. MIL assumes each annotated region as a “bag” containing one or more instances of a class. If at least a single instance of a class is matched inside a bag, it is treated as a true positive. If no instances of the class are detected inside the bag, then the entire bag is treated as a false negative. In case an instance of class is mismatched inside a bag belonging to some other behavior, then it is treated as a false positive. Finally, if no mismatch occurs inside a bag of a non-relevant class, then the entire bag is treated as a true negative.

Next, we investigate the utility of our algorithm for separating distinct chicken behaviors. A healthy chicken is expected to display a set of behaviors \cite{banerjee2012remote}. In this work, we only considered pecking, preening and dustbathing behaviors. Pecking is perhaps the most familiar behavior in chickens which may be performed tens of thousands of times each day.

\begin{figure}[h]
\centering
\includegraphics[width=.8\textwidth]{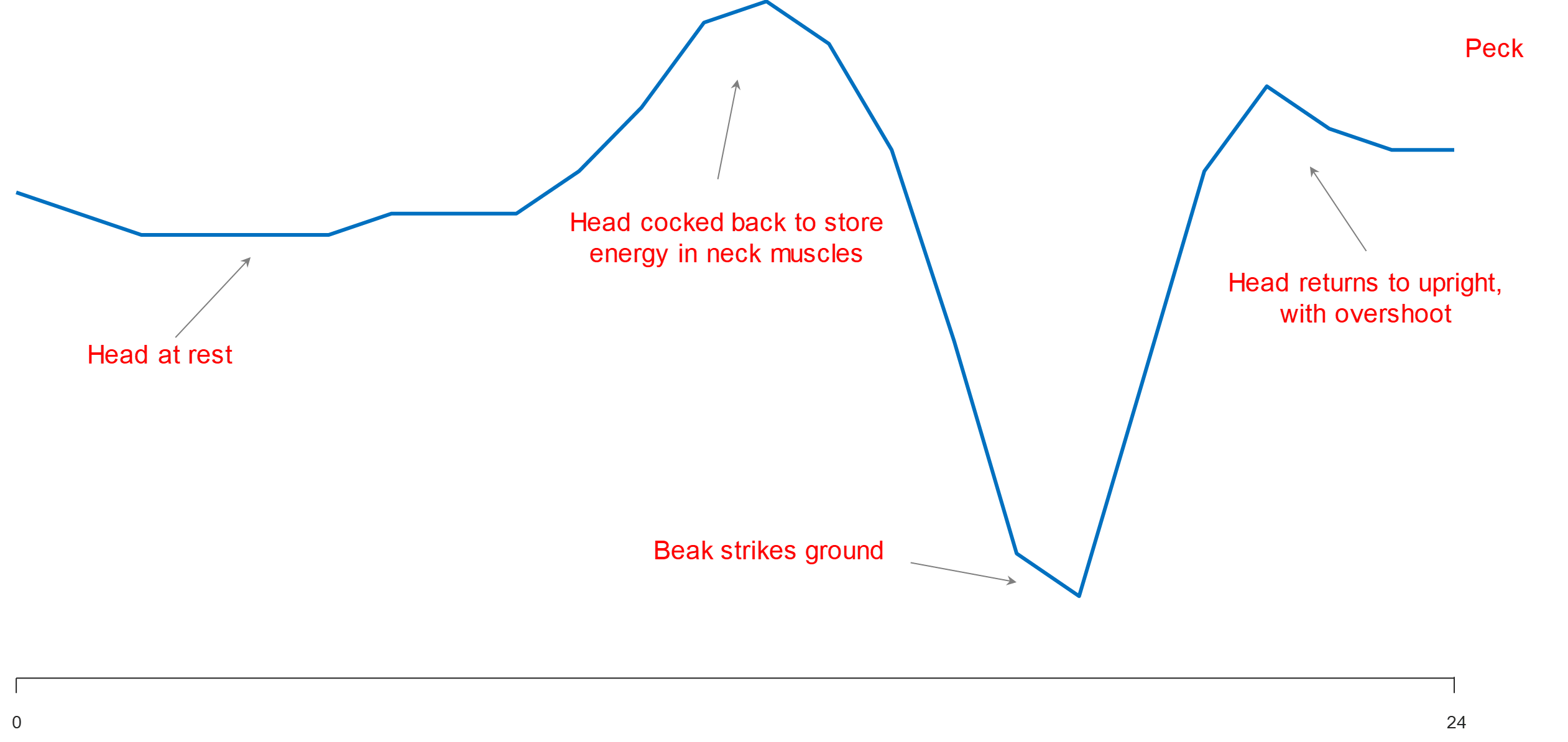}
\caption{The shape of a chicken peck, used as the query for creating a shape-based model for classification of pecking behavior.}~\label{fig:kdd_8}
\end{figure}

\begin{figure}[h]
\centering
\includegraphics[width=\textwidth]{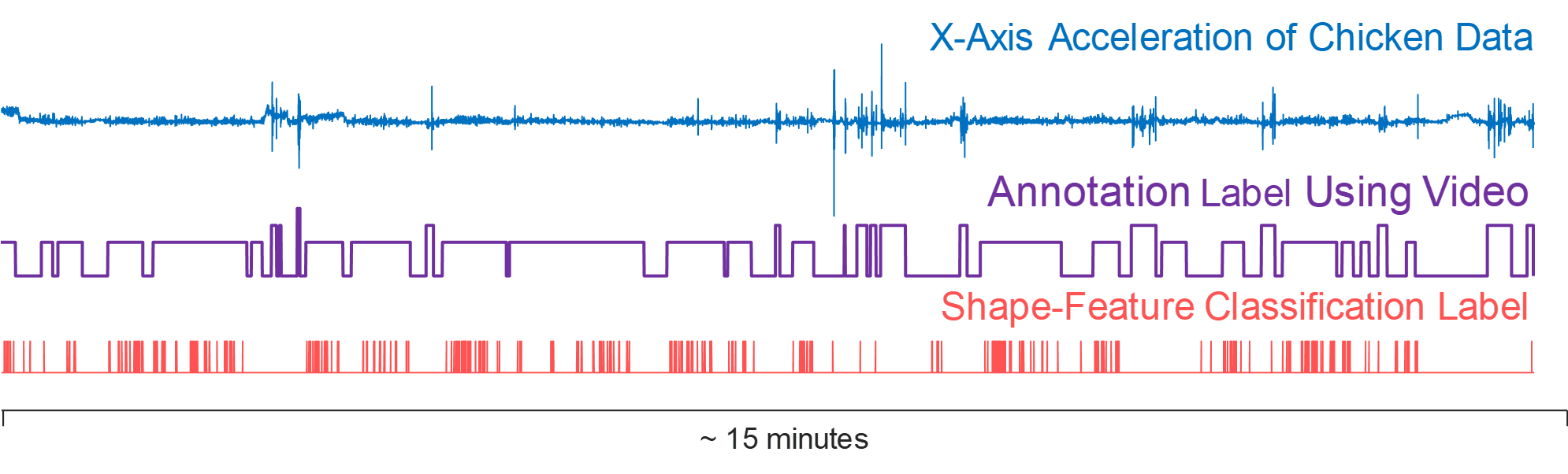}
\caption{The shape-feature classification for the pecking behavior in rose pink. For clarity only the X-Axis is shown in blue with the annotation video from the video in purple.}~\label{fig:kdd_9}
\end{figure}

Figure \ref{fig:kdd_8} shows the typical query-template we used for creating the shape-based model for the pecking behavior. Given the query in Figure 8, we calculate the distance profile (see definition 4) for the training dataset. Then, we calculate feature vectors and separate pecking and non-pecking instances based on calculated distances and feature vectors with a high degree of confidence. The classification of training dataset for pecking behavior based on shape-feature algorithm is shown in Figure \ref{fig:kdd_9}. Table \ref{table:kdd_pecking} presents the confusion matrix for the classification of pecking behavior in the test dataset.

\begin{table}[h]
\caption{Confusion matrix for pecking behavior}
\begin{center}
\begin{tabular}{|c|c|c|}
\hline
Classes                & Pecking  & Non-Pecking \\ \hline
Pecking        & \textbf{25 (TP)}  & \textbf{4 (FN)}  \\ \hline
Non-Pecking   & \textbf{2 (FP)}  & \textbf{40 (TN)} \\ \hline
\end{tabular}
\end{center}
\label{table:kdd_pecking}
\end{table}
\vspace{-0.25in}
\[
    Accuracy = \frac{TP + TN}{TP + TN + FP + FN} = \textbf{86 \%}
\]

\[
    Precision = \frac{TP}{TP + FP} = \textbf{92 \%}
\]

\[
    Recall = \frac{TP}{TP + FN} = \textbf{91 \%}
\]

Given the results above, our classification model has 86\% precision and 92\% recall in matching instances of the pecking behavior. Overall, the classifier has 91\% accuracy for the preening behavior, which compares very favorably to 70\% default rate (i.e., guessing every observed object as the majority class).

\begin{figure}[h]
\centering
\includegraphics[width=\textwidth]{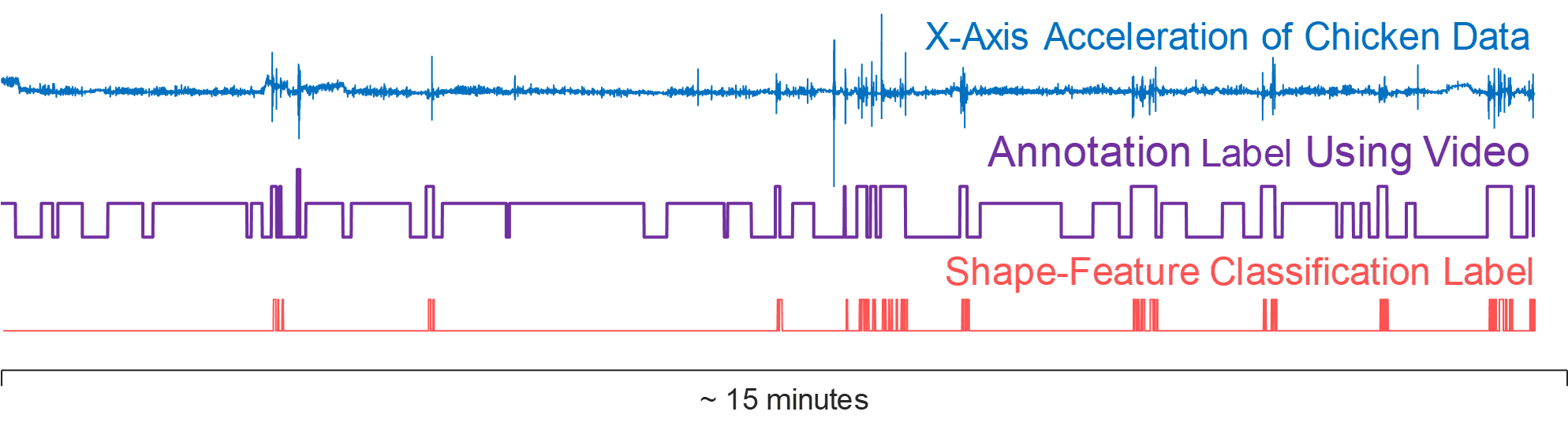}
\caption{The shape-feature classification for the preening behavior in rose pink. For clarity only the X-Axis is shown in blue with the annotation video from the video in purple.}~\label{fig:kdd_10}
\end{figure}

The preening behavior does not have a well-conserved shape. Thus, we should turn our attention to features. Similarly, we calculate the complexity feature vectors for the training dataset. Next, based on complexity feature vector we separate preening/dustbathing from non-class (i.e. pecking and noisy data) instances. Finally, we use spectral density feature vector to differentiate between preening and non-preening (i.e. dustbathing) instances as shown in Figure \ref{fig:kdd_10} and Figure \ref{fig:kdd_11}. Table \ref{table:kdd_preening} presents the confusion matrix for the classification of preening behavior in the test dataset.

\vspace{-0.25in}

\begin{table}[h]
\caption{Confusion matrix for preening behavior}
\begin{center}
\begin{tabular}{|c|c|c|}
\hline
Classes                & Preening  & Non-Preening \\ \hline
Preening        & \textbf{14 (TP)}  & \textbf{0 (FN)}  \\ \hline
Non-Preening   & \textbf{0 (FP)}  & \textbf{57 (TN)} \\ \hline
\end{tabular}
\end{center}
\label{table:kdd_preening}
\end{table}

\[
    Accuracy = \frac{TP + TN}{TP + TN + FP + FN} = \textbf{100 \%}
\]

\[
    Precision = \frac{TP}{TP + FP} = \textbf{100 \%}
\]

\[
    Recall = \frac{TP}{TP + FN} = \textbf{100 \%}
\]

Given the results above, our classification model has a 100\% precision and recall in matching instances of the preening behavior. Overall, the classifier has 100\% accuracy for the preening behavior, which compares very favorably to 70\% default rate.

\begin{figure}[h]
\centering
\includegraphics[width=\textwidth]{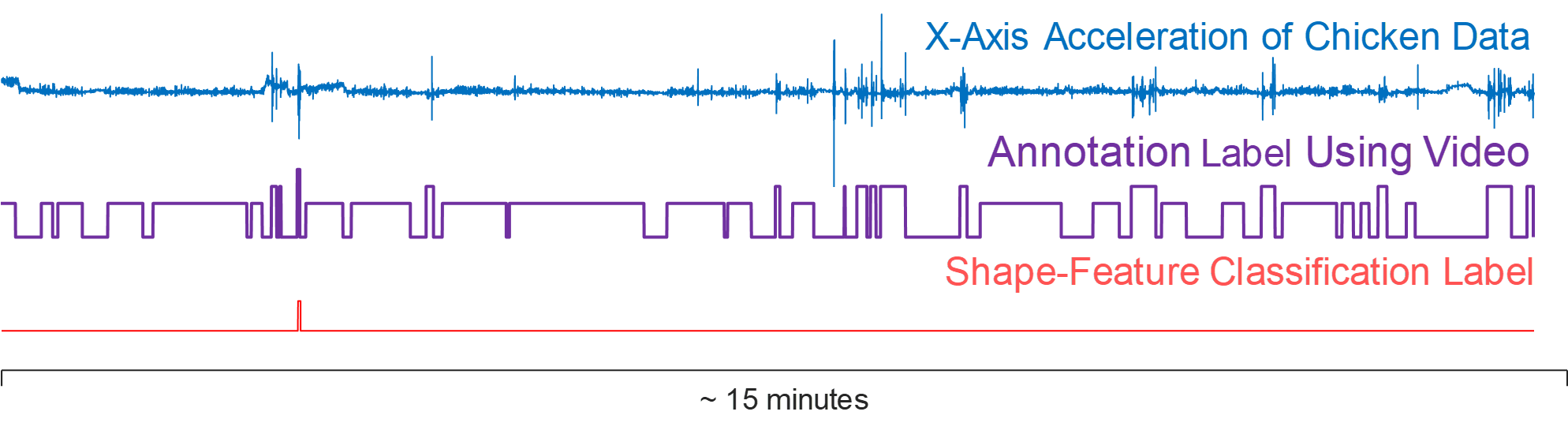}
\caption{The shape-feature classification for the preening behavior in rose pink. For clarity only the X-Axis is shown in blue with the annotation video from the video in purple.}~\label{fig:kdd_11}
\end{figure}

\begin{table}[h]
\caption{Confusion matrix for dustbathing behavior}
\begin{center}
\begin{tabular}{|c|c|c|}
\hline
Classes                & Dustbathing  & Non-Dustbathing \\ \hline
Dustbathing        & \textbf{1 (TP)}  & \textbf{0 (FN)}  \\ \hline
Non-Dustbathing   & \textbf{0 (FP)}  & \textbf{70 (TN)} \\ \hline
\end{tabular}
\end{center}
\label{table:kdd_dustbathing}
\end{table}

\[
    Accuracy = \frac{TP + TN}{TP + TN + FP + FN} = \textbf{100 \%}
\]

\[
    Precision = \frac{TP}{TP + FP} = \textbf{100 \%}
\]

\[
    Recall = \frac{TP}{TP + FN} = \textbf{100 \%}
\]

Table \ref{table:kdd_dustbathing} presents the confusion matrix for the classification of dustbathing behavior in the test dataset. Given these evaluation results, our model has 1.00 precision in matching dustbathing subsequences and 1.00 recall in matching relevant instances of the dustbathing behavior. Finally, the model has 100\% overall accuracy in matching dustbathing subsequences compared to 99\% default rate (i.e., guessing every observed object as the majority class).

\begin{figure}[h]
\centering
\includegraphics[width=\textwidth]{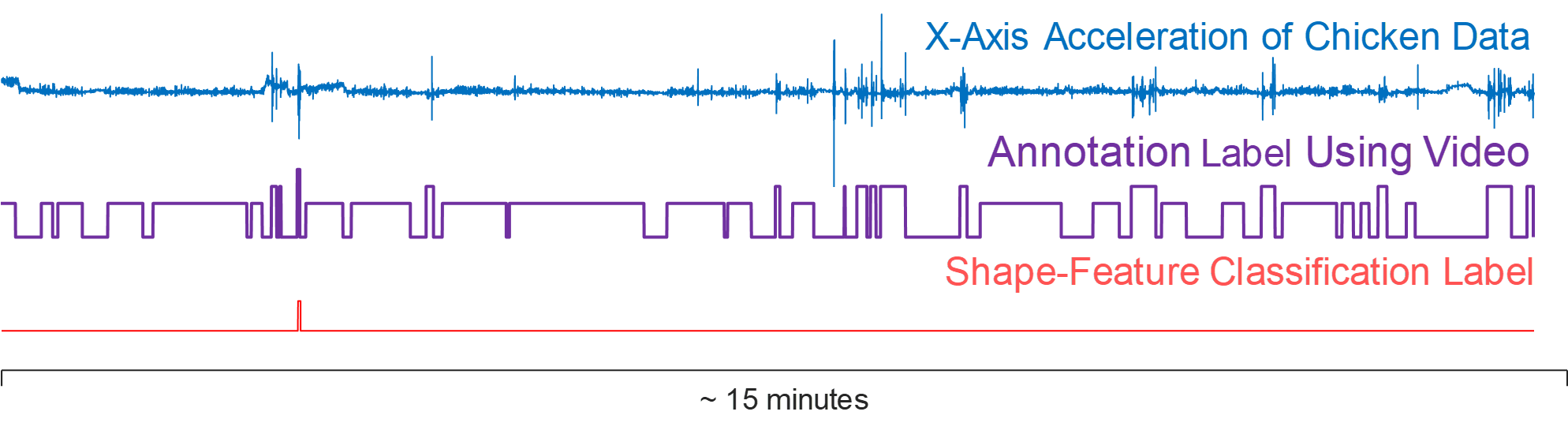}
\caption{The shape-feature classification for the preening behavior in rose pink. For clarity only the X-Axis is shown in blue with the annotation video from the video in purple.}~\label{fig:kdd_11}
\end{figure}

\subsection{A Critical Lesion Study}

Our fundamental claim in this work is that using a combined shape/feature modality is superior in some domains, including the domain-at-hand. There is a very simple way to demonstrate this. We can repeat the previous experiments twice, once using only shape, and once using only features. Everything else remains exactly the same. Table \ref{table:kdd_comparison} shows the results, which confirm the superiority of our more expressive approach.

While dustbathing is easy to recognize with either modality, a feature-based approach has difficulty with pecking, and a shape based-approach has difficulty with preening. As the only difference in these three experiments are the allowed modalities, thus we can attribute our success to more general algorithm.

\begin{table}[h]
\caption{Shape-feature classification for chicken behaviors}
\begin{center}
\begin{tabular}{|c|c|ccc|}
\hline
Classification                 & Activity    & \multicolumn{1}{c|}{Precision} & \multicolumn{1}{c|}{Recall} & Accuracy \\ \hline
\multirow{3}{*}{Shape}         & Pecking     & 0.71                           & 0.81                        & 0.85     \\ 
                               & Preening    & 0.91                           & 0.71                        & 0.93     \\
                               & Dustbathing & 1                              & 1                           & 1        \\ \hline
\multirow{3}{*}{Feature}       & Pecking     & 0.60                           & 0.77                        & 0.77     \\
                               & Preening    & 1                              & 1                           & 1        \\
                               & Dustbathing & 1                              & 1                           & 1        \\ \hline
\multirow{3}{*}{Shape-Feature} & Pecking     & 0.86                           & 0.92                        & 0.91     \\ 
                               & Preening    & 1                              & 1                           & 1        \\
                               & Dustbathing & 1                              & 1                           & 1        \\ \hline
\end{tabular}
\end{center}
\label{table:kdd_comparison}
\end{table}

\subsection{Case Study: A Day of Chicken Life}

Up to now, we trained and tested the proposed shape-feature based classification algorithm on labeled data. However, as we mentioned early in this study, we believe that the lessons learned with the short (labeled) dataset can be generalized to other chickens that were not previously seen.

\begin{figure}[h]
\centering
\includegraphics[width=\textwidth]{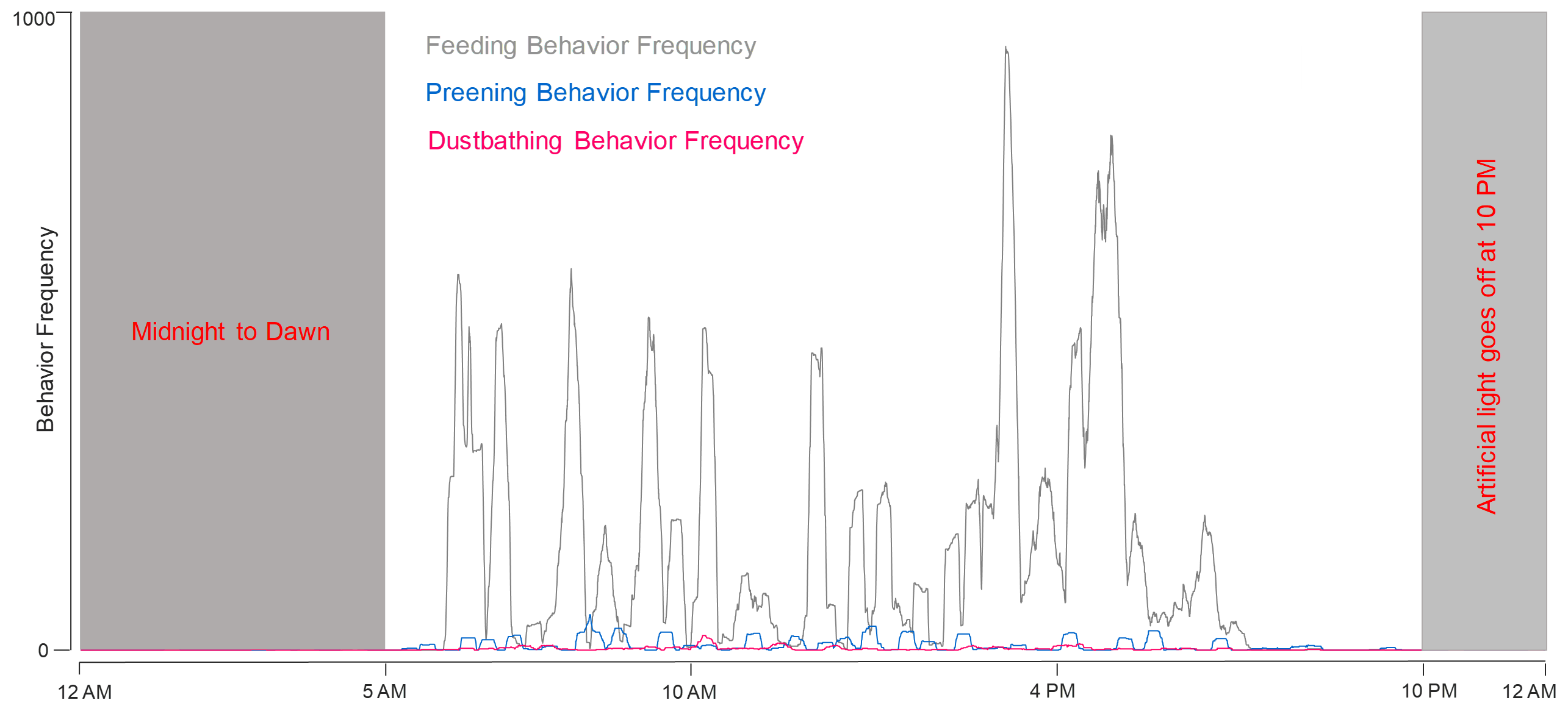}
\caption{Shape-Feature based classification of a 24-hour of chicken data. The gray regions denote nighttime.}~\label{fig:kdd_12}
\end{figure}

We start by looking into classifying a day of data for a single chicken, as shown in Figure \ref{fig:kdd_12}. Most animals have a daily recurrent pattern of activity called a “Circadian Rhythm”. We examine the existence of such pattern in Figure \ref{fig:kdd_12}. As can be seen in the figure, there are not many activities before sunrise and after artificial lights went off at 10 pm.

This makes sense as chickens are visual creatures and quickly fall to sleep with a lack of light. Next, we go one step further and we put the classification results into work to differentiate between healthy and sick chickens based on the frequency (count) of the behaviors.

\subsection{Case Study: Healthy vs. Sick Chickens}

Diseases are of particular importance due to the heavy economic losses they cause in poultry production \cite{murillo2020parasitic}. Rapid detection and diagnosis allow for decreasing the costs associated with the disease. In this section, we will use our feature-shape algorithm to recognize unhealthy chickens to combat the spread of diseases in poultry.

We look into two 24-hour days of chicken data for all the chickens. Given the field-inspection and manual verification by entomologist researchers, we know that the chickens were all healthy on the first day and all were sick (infested with ectoparasites) on the other day. The aim is that the shape-feature classification results can help to distinguish between healthy and sick chickens with an acceptable confidence.

\begin{figure}[!b]
\centering
\includegraphics[width=\textwidth]{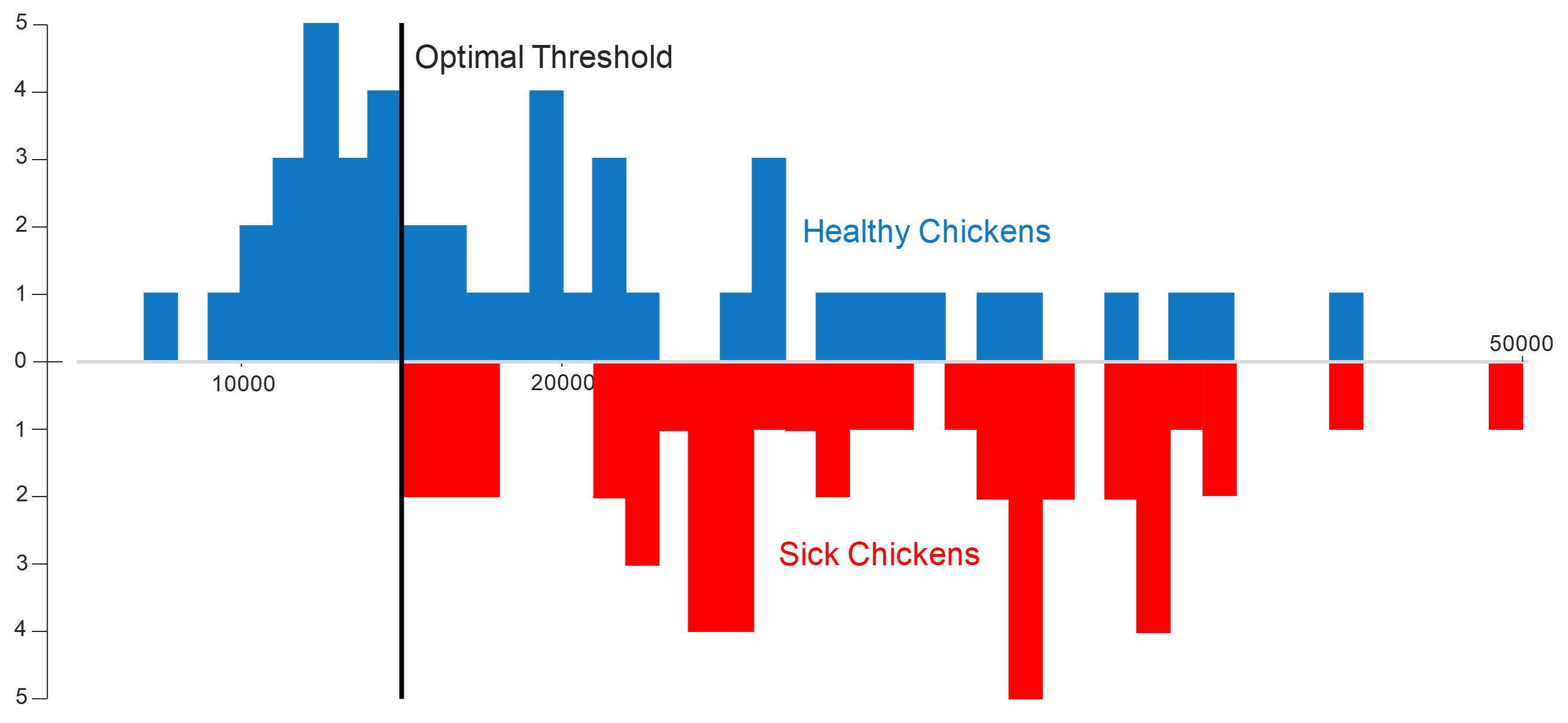}
\caption{The histogram for distribution of the number of pecking behaviors in healthy (blue) and sick (red) chickens.}~\label{fig:kdd_13}
\end{figure}

We start by looking at the histogram for the distribution of the number of pecking behaviors for all the chickens when they were healthy (blue) and sick (red), as shown in Figure \ref{fig:kdd_13}. As evident in the figure the peck counts for healthy and sick chickens have some overlaps.

\begin{figure}[h]
\centering
\includegraphics[width=\textwidth]{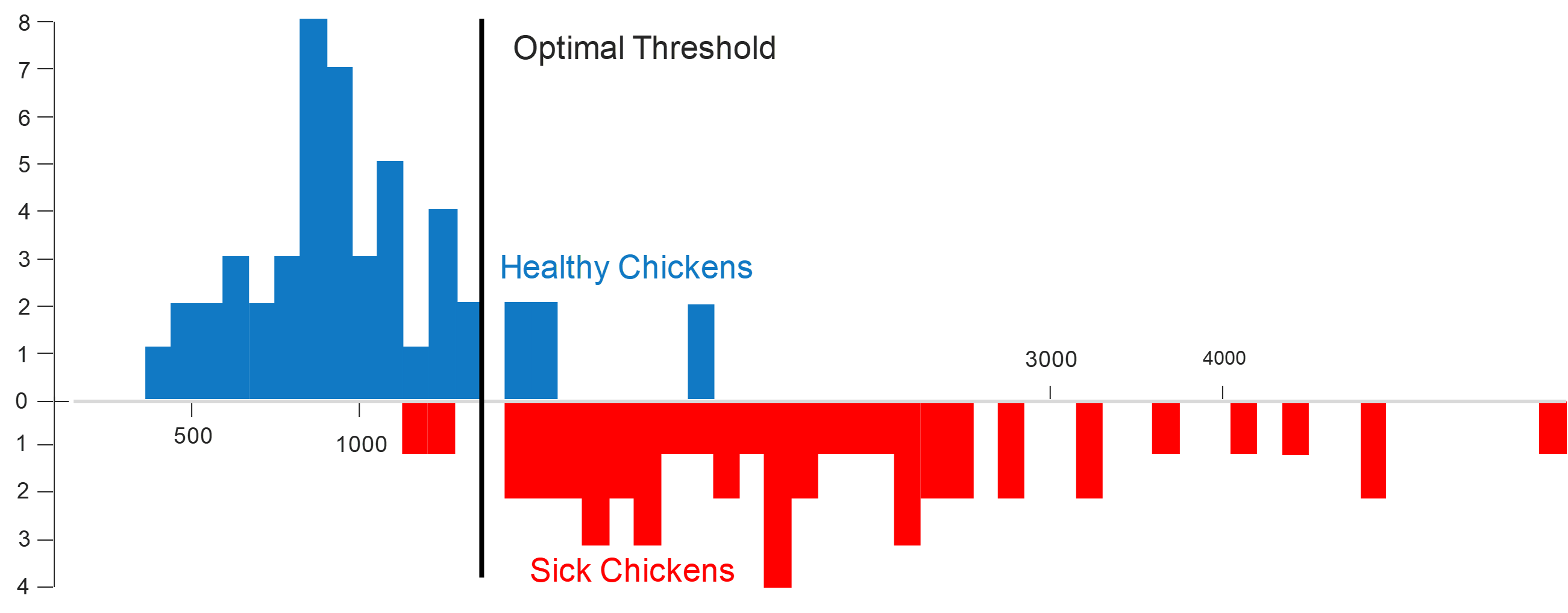}
\caption{The histogram for distribution of the number of preening behaviors in healthy (blue) and sick (red) chickens.}~\label{fig:kdd_14}
\end{figure}

Next, we proceed to inspecting the distribution of the number of preening behaviors for all the chickens when they are healthy (blue) and sick (red). As shown in Figure \ref{fig:kdd_14} the distribution of preening counts for healthy chickens is skewed to the left of the histogram while the distribution of the preening counts for sick chickens is concentrated at the center and right side of the histogram. This shows that the shape-feature classification algorithm can distinguish between sick and healthy chickens based on the preening count with a good confidence.

\begin{figure}[t]
\centering
\includegraphics[width=\textwidth]{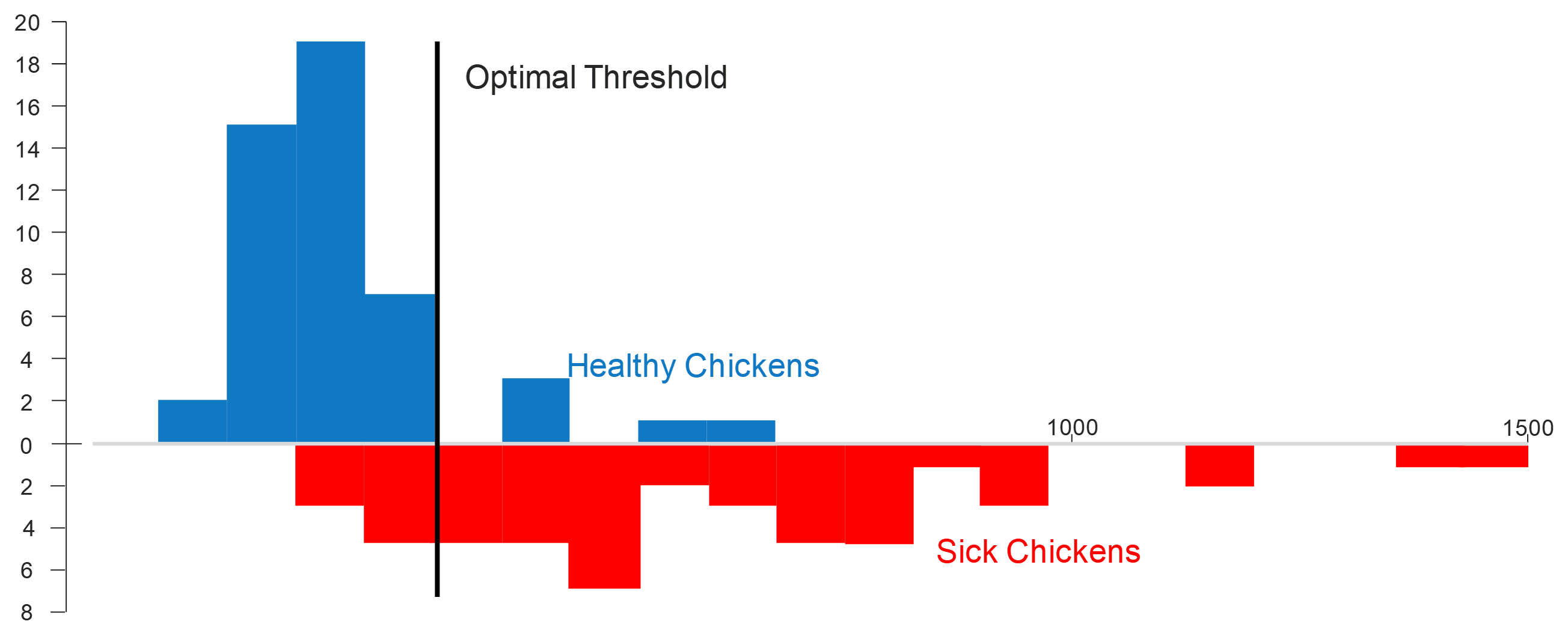}
\caption{The histogram for distribution of the number of dustbathing behaviors in healthy (blue) and sick (red) chickens.}~\label{fig:kdd_15}
\end{figure}

Finally, in Figure \ref{fig:kdd_15}, we look into the distribution of the number of dustbathing behaviors for healthy (blue) and sick (red) chickens. It is obvious that the distribution of dustbathing for healthy chickens is mostly accumulated at the left side of the histogram. Whereas the distribution of the dustbathing count for sick chickens is spread over at the center of the histogram.

In overall, the pecking behavior seems to provide an acceptable  confidence for distinguishing healthy and sick chickens. The preening and dustbathing behaviors show great confidence towards differentiating between sick and healthy chickens.

It is fair to say that our proposed method is expressive since the only difference between our algorithm and the other two methods (i.e. shape-based classification and feature-based classification) is the way we combined those two possibilities. Nothing else has changed. Therefore, we can attribute any success only to the increased expressiveness of the proposed method.

\section{Conclusion}

we introduced an algorithm to classify behaviors, using both shape and feature measures, in weakly labeled time series data. We demonstrated, with an extensive empirical study, that our algorithm can robustly classify real, noisy and complex datasets, based on a combination of shape and features, and tested our proposed algorithm on real-world datasets. Our ideas are currently been evaluated with a large-scale field trial involving
hundreds of birds. While our study was motivated by a pressing problem in poultry welfare, it could clearly be used in other realworld data problems. For example, oil and gas production is telemetry-rich endeavor. Moreover, it features all the elements we mentioned that make our poultry problem so difficult. In particular:

\begin{itemize}
    \item Data is often only weakly labeled. For example, the ultimate
Key Performance Indicator (KPI) may only be obtained once a shift by taking a physical sample and measuring some metric. If it is bad, we get only a weak label such as “sometime in the last eight hours something bad happened”.

\item The duration of different classes can be at different scales, for
example foaming might happen in minutes, whereas clogging
may take hours or longer.

\item Different classes may best manifest themselves in just one of the shape or feature modalities. For example, foaming is typically recognized (at least by eye) in feature space, whereas sticking-valve is recognized in shape (as a “step pattern”).
\end{itemize}

Thus, we believe that our observations and algorithms may be more generally applicable as researchers move beyond the toy problems used in the UCR archive.
\chapter{Conclusions}

\newpage

In this dissertation we demonstrated that neither of the shape-based or feature-based approaches will be perfect for all classification problems. On a given problem, it might be possible that shape classification is better for some subset of the behaviors, and the feature classification technique is better for other subset of behaviors.

Next, we introduced the hybrid shape-feature classification algorithm to classify behaviors, using both shape and feature measures, in weakly labeled time series data collected from sensors to quantify specific behaviors performed by the subject. We demonstrate that our algorithm can robustly classify real, noisy, and complex datasets, based on a combination of shape and features, and tested our proposed algorithm on real-world datasets.

As the main case-study of this dissertation, we applied the proposed shape-feature classification approach on chicken data. The course of this dissertation was in the following order:

\begin{itemize}
    \item Chapter 1 briefly discussed abundance and technological advances in sensing technologies, which in turn allowed for mass volume data collection. Given the capability to collect excessively large amounts of data, the importance of efficient and scalable time series data mining algorithms was emphasized.
    
    \item Chapter 2 explained the data engineering efforts towards setup, installation and collection of data from data-logger and real-time sensors. Next, the efforts were focused on transferring collected data to the cloud storage and subsequently data cleaning and preparation tasks were discussed.
    
    \item Chapter 3 presented shape-based classification which utilized euclidean distance measure to look for similar subsequences towards the task of time series classification.
    
    \item Chapter 4 was dedicated to the novel hybrid shape-feature classification approach. The superiority of the shape-feature classification approach lies in the fact the some behaviors (e.g., smooth and repetitive behaviors) are better classified based on their shape, while other behaviors (e.g., complex and fast behaviors) are better classified based on various mathematical and statistical features. Next, we applied the shape-feature classification approach on terabytes of chicken data and suggested correlation between timing and frequency of behaviors and the overall health status of the chickens.
    
    \item Next, the utility of the proposed shape-feature classification approach was demonstrated on human activity data, to emphasize wide range of data on which the shape-feature approach could be employed.
\end{itemize}

\nocite{*}
\bibliographystyle{plain}
\bibliography{thesis}

\begin{thebibliography}{10}

\bibitem{abdoli2020fitbit}
Alireza Abdoli, Sara Alaee, Shima Imani, Amy Murillo, Alec Gerry, Leslie
  Hickle, and Eamonn Keogh.
\newblock Fitbit for chickens? time series data mining can increase the
  productivity of poultry farms.
\newblock In {\em Proceedings of the 26th ACM SIGKDD International Conference
  on Knowledge Discovery \& Data Mining}, pages 3328--3336, 2020.

\bibitem{abdoli2018stationary}
Alireza Abdoli and Philip Brisk.
\newblock Stationary-mixing field-programmable pin-constrained digital
  microfluidic biochip.
\newblock {\em Microelectronics journal}, 77:34--48, 2018.

\bibitem{abdoli2015field}
Alireza Abdoli and Ali Jahanian.
\newblock Field-programmable cell array pin-constrained digital microfluidic
  biochip.
\newblock In {\em 2015 22nd Iranian Conference on Biomedical Engineering
  (ICBME)}, pages 48--53. IEEE, 2015.

\bibitem{abdoli2015general}
Alireza Abdoli and Ali Jahanian.
\newblock A general-purpose field-programmable pin-constrained digital
  microfluidic biochip.
\newblock In {\em 2015 18th CSI International Symposium on Computer
  Architecture and Digital Systems (CADS)}, pages 1--6. IEEE, 2015.

\bibitem{abdoli2020cost}
Alireza Abdoli and Ali Jahanian.
\newblock A cost \& performance-efficient field-programmable pin-constrained
  digital microfluidic biochip.
\newblock {\em arXiv preprint arXiv:2008.09975}, 2020.

\bibitem{abdoli2019time}
Alireza Abdoli, Amy~C Murillo, Alec~C Gerry, and Eamonn~J Keogh.
\newblock Time series classification: Lessons learned in the (literal) field
  while studying chicken behavior.
\newblock In {\em 2019 IEEE International Conference on Big Data (Big Data)},
  pages 5962--5964. IEEE, 2019.

\bibitem{abdoli2018time}
Alireza Abdoli, Amy~C Murillo, Chin-Chia~M Yeh, Alec~C Gerry, and Eamonn~J
  Keogh.
\newblock Time series classification to improve poultry welfare.
\newblock In {\em 2018 17TH IEEE International conference on machine learning
  and applications (ICMLA)}, pages 635--642. IEEE, 2018.

\bibitem{alaee2020features}
Sara Alaee, Alireza Abdoli, Christian Shelton, Amy~C Murillo, Alec~C Gerry, and
  Eamonn Keogh.
\newblock Features or shape? tackling the false dichotomy of time series
  classification∗.
\newblock In {\em Proceedings of the 2020 SIAM International Conference on Data
  Mining}, pages 442--450. SIAM, 2020.

\bibitem{amores2013multiple}
Jaume Amores.
\newblock Multiple instance classification: Review, taxonomy and comparative
  study.
\newblock {\em Artificial intelligence}, 201:81--105, 2013.

\bibitem{banerjee2012remote}
Debasmit Banerjee, Subir Biswas, Courtney Daigle, and Janice~M Siegford.
\newblock Remote activity classification of hens using wireless body mounted
  sensors.
\newblock In {\em 2012 Ninth International Conference on Wearable and
  Implantable Body Sensor Networks}, pages 107--112. IEEE, 2012.

\bibitem{barwick2018categorising}
Jamie Barwick, David~W Lamb, Robin Dobos, Mitchell Welch, and Mark Trotter.
\newblock Categorising sheep activity using a tri-axial accelerometer.
\newblock {\em Computers and Electronics in Agriculture}, 145:289--297, 2018.

\bibitem{batista2011sigkdd}
Gustavo~E Batista, Eamonn~J Keogh, Agenor Mafra-Neto, and Edgar Rowton.
\newblock Sigkdd demo: sensors and software to allow computational entomology,
  an emerging application of data mining.
\newblock In {\em Proceedings of the 17th ACM SIGKDD international conference
  on Knowledge discovery and data mining}, pages 761--764, 2011.

\bibitem{batista2014cid}
Gustavo~EAPA Batista, Eamonn~J Keogh, Oben~Moses Tataw, and Vinicius~MA
  De~Souza.
\newblock Cid: an efficient complexity-invariant distance for time series.
\newblock {\em Data Mining and Knowledge Discovery}, 28(3):634--669, 2014.

\bibitem{blatchford2017animal}
RA~Blatchford.
\newblock Animal behavior and well-being symposium: Poultry welfare
  assessments: Current use and limitations.
\newblock {\em Journal of animal science}, 95(3):1382--1387, 2017.

\bibitem{brown1974effect}
N~Sandra Brown.
\newblock The effect of louse infestation, wet feathers, and relative humidity
  on the grooming behavior of the domestic chicken.
\newblock {\em Poultry Science}, 53(5):1717--1719, 1974.

\bibitem{chiu2003probabilistic}
Bill Chiu, Eamonn Keogh, and Stefano Lonardi.
\newblock Probabilistic discovery of time series motifs.
\newblock In {\em Proceedings of the ninth ACM SIGKDD international conference
  on Knowledge discovery and data mining}, pages 493--498, 2003.

\bibitem{da2017clustering}
Josenildo~Costa da~Silva, Gustavo~HBS Oliveira, Stefano Lodi, and Matthias
  Klusch.
\newblock Clustering distributed short time series with dense patterns.
\newblock In {\em 2017 16th IEEE International Conference on Machine Learning
  and Applications (ICMLA)}, pages 34--41. IEEE, 2017.

\bibitem{daigle2012noncaged}
CL~Daigle, D~Banerjee, S~Biswas, and JM~Siegford.
\newblock Noncaged laying hens remain unflappable while wearing body-mounted
  sensors: Levels of agonistic behaviors remain unchanged and resource use is
  not reduced after habituation.
\newblock {\em Poultry Science}, 91(10):2415--2423, 2012.

\bibitem{daigle2014moving}
Courtney~L Daigle, Debasmit Banerjee, Robert~A Montgomery, Subir Biswas, and
  Janice~M Siegford.
\newblock Moving gis research indoors: Spatiotemporal analysis of agricultural
  animals.
\newblock {\em PLoS One}, 9(8):e104002, 2014.

\bibitem{dau2017matrix}
Hoang~Anh Dau and Eamonn Keogh.
\newblock Matrix profile v: A generic technique to incorporate domain knowledge
  into motif discovery.
\newblock In {\em Proceedings of the 23rd ACM SIGKDD international conference
  on knowledge discovery and data mining}, pages 125--134, 2017.

\bibitem{dawkins2004chicken}
Marian~Stamp Dawkins, Christl~A Donnelly, and Tracey~A Jones.
\newblock Chicken welfare is influenced more by housing conditions than by
  stocking density.
\newblock {\em Nature}, 427(6972):342--344, 2004.

\bibitem{ding2008querying}
Hui Ding, Goce Trajcevski, Peter Scheuermann, Xiaoyue Wang, and Eamonn Keogh.
\newblock Querying and mining of time series data: experimental comparison of
  representations and distance measures.
\newblock {\em Proceedings of the VLDB Endowment}, 1(2):1542--1552, 2008.

\bibitem{doppler2009variability}
Jakob Doppler, Gerald Holl, Alois Ferscha, Marquart Franz, Cornel Klein, Marcos
  dos Santos~Rocha, and Andreas Zeidler.
\newblock Variability in foot-worn sensor placement for activity recognition.
\newblock In {\em 2009 International Symposium on Wearable Computers}, pages
  143--144. IEEE, 2009.

\bibitem{ELAN}
ELAN.
\newblock Global poultry trends - developing countries main drivers in chicken
  consumption.

\bibitem{everts2016information}
Sarah Everts.
\newblock Information overload.
\newblock {\em Science History Institute. July}, 18:2016, 2016.

\bibitem{FAO_Housing}
FAO.
\newblock Oecd-fao agricultural.

\bibitem{fulcher2013highly}
Ben~D Fulcher, Max~A Little, and Nick~S Jones.
\newblock Highly comparative time-series analysis: the empirical structure of
  time series and their methods.
\newblock {\em Journal of the Royal Society Interface}, 10(83):20130048, 2013.

\bibitem{gao2017iterative}
Yifeng Gao, Jessica Lin, and Huzefa Rangwala.
\newblock Iterative grammar-based framework for discovering variable-length
  time series motifs.
\newblock In {\em 2017 IEEE International Conference on Data Mining (ICDM)},
  pages 111--116. IEEE, 2017.

\bibitem{gerry2019promoting}
AC~Gerry, AC~Murillo, et~al.
\newblock Promoting biosecurity through insect management at animal facilities.
\newblock {\em Biosecurity in animal production and veterinary medicine: from
  principles to practice}, pages 243--281, 2019.

\bibitem{gharghabi2017matrix}
Shaghayegh Gharghabi, Yifei Ding, Chin-Chia~Michael Yeh, Kaveh Kamgar, Liudmila
  Ulanova, and Eamonn Keogh.
\newblock Matrix profile viii: domain agnostic online semantic segmentation at
  superhuman performance levels.
\newblock In {\em 2017 IEEE international conference on data mining (ICDM)},
  pages 117--126. IEEE, 2017.

\bibitem{gharghabi2018matrix}
Shaghayegh Gharghabi, Shima Imani, Anthony Bagnall, Amirali Darvishzadeh, and
  Eamonn Keogh.
\newblock Matrix profile xii: Mpdist: a novel time series distance measure to
  allow data mining in more challenging scenarios.
\newblock In {\em 2018 IEEE International Conference on Data Mining (ICDM)},
  pages 965--970. IEEE, 2018.

\bibitem{gharghabi2020ultra}
Shaghayegh Gharghabi, Shima Imani, Anthony Bagnall, Amirali Darvishzadeh, and
  Eamonn Keogh.
\newblock An ultra-fast time series distance measure to allow data mining in
  more complex real-world deployments.
\newblock {\em Data Mining and Knowledge Discovery}, 34:1104--1135, 2020.

\bibitem{giusti2016improved}
Rafael Giusti, Diego~F Silva, and Gustavo~EAPA Batista.
\newblock Improved time series classification with representation diversity and
  svm.
\newblock In {\em 2016 15th IEEE International Conference on Machine Learning
  and Applications (ICMLA)}, pages 1--6. IEEE, 2016.

\bibitem{imani2021multi}
Shima Imani, Alireza Abdoli, Ali Beyram, Azam Imani, and Eamonn Keogh.
\newblock Multi-window-finder: Domain agnostic window size for time series
  data.
\newblock 2021.

\bibitem{imanitime2cluster}
Shima Imani, Alireza Abdoli, and Eamonn Keogh.
\newblock Time2cluster: Clustering time series using neighbor information.

\bibitem{imani2019putting}
Shima Imani, Sara Alaee, and Eamonn Keogh.
\newblock Putting the human in the time series analytics loop.
\newblock In {\em Companion proceedings of the 2019 World Wide Web conference},
  pages 635--644, 2019.

\bibitem{imani2019matrix}
Shima Imani and Eamonn Keogh.
\newblock Matrix profile xix: time series semantic motifs: a new primitive for
  finding higher-level structure in time series.
\newblock In {\em 2019 IEEE International Conference on Data Mining (ICDM)},
  pages 329--338. IEEE, 2019.

\bibitem{imani2018matrix}
Shima Imani, Frank Madrid, Wei Ding, Scott Crouter, and Eamonn Keogh.
\newblock Matrix profile xiii: Time series snippets: a new primitive for time
  series data mining.
\newblock In {\em 2018 IEEE international conference on big knowledge (ICBK)},
  pages 382--389. IEEE, 2018.

\bibitem{karimi2019border}
Mohsen Karimi, Ali Jahanshahi, Abbas Mazloumi, and Hadi~Zamani Sabzi.
\newblock Border gateway protocol anomaly detection using neural network.
\newblock In {\em 2019 IEEE International Conference on Big Data (Big Data)},
  pages 6092--6094. IEEE, 2019.

\bibitem{kreil2016coping}
Matthias Kreil, Bernhard Sick, and Paul Lukowicz.
\newblock Coping with variability in motion based activity recognition.
\newblock In {\em Proceedings of the 3rd International Workshop on Sensor-based
  Activity Recognition and Interaction}, pages 1--8, 2016.

\bibitem{lara2012survey}
Oscar~D Lara and Miguel~A Labrador.
\newblock A survey on human activity recognition using wearable sensors.
\newblock {\em IEEE communications surveys \& tutorials}, 15(3):1192--1209,
  2012.

\bibitem{lee2017anomaly}
Doyup Lee.
\newblock Anomaly detection in multivariate non-stationary time series for
  automatic dbms diagnosis.
\newblock In {\em 2017 16th IEEE International Conference on Machine Learning
  and Applications (ICMLA)}, pages 412--419. IEEE, 2017.

\bibitem{li2020automated}
N~Li, Z~Ren, D~Li, and L~Zeng.
\newblock Automated techniques for monitoring the behaviour and welfare of
  broilers and laying hens: towards the goal of precision livestock farming.
\newblock {\em animal}, 14(3):617--625, 2020.

\bibitem{lubba2019catch22}
Carl~H Lubba, Sarab~S Sethi, Philip Knaute, Simon~R Schultz, Ben~D Fulcher, and
  Nick~S Jones.
\newblock catch22: Canonical time-series characteristics.
\newblock {\em Data Mining and Knowledge Discovery}, 33(6):1821--1852, 2019.

\bibitem{mcfarland2013assessing}
Richard McFarland, Robyn~S Hetem, Andrea Fuller, Duncan Mitchell, S~Peter
  Henzi, and Louise Barrett.
\newblock Assessing the reliability of biologger techniques to measure activity
  in a free-ranging primate.
\newblock {\em Animal Behaviour}, 4(85):861--866, 2013.

\bibitem{moreau2009use}
Maeg Moreau, Stefan Siebert, Andreas Buerkert, and Eva Schlecht.
\newblock Use of a tri-axial accelerometer for automated recording and
  classification of goats’ grazing behaviour.
\newblock {\em Applied animal behaviour science}, 119(3-4):158--170, 2009.

\bibitem{mueen2017fastest}
Abdullah Mueen, Krishnamurthy Viswanathan, Chetan Gupta, and Eamonn Keogh.
\newblock The fastest similarity search algorithm for time series subsequences
  under euclidean distance.
\newblock {\em url: www. cs. unm. edu/\~{} mueen/FastestSimilaritySearch. html
  (accessed 24 May, 2016)}, 2017.

\bibitem{murillo2020parasitic}
Amy~C Murillo, Alireza Abdoli, Richard~A Blatchford, Eamonn~J Keogh, and Alec~C
  Gerry.
\newblock Parasitic mites alter chicken behaviour and negatively impact animal
  welfare.
\newblock {\em Scientific reports}, 10(1):1--12, 2020.

\bibitem{murillo2016sulfur}
Amy~C Murillo and Bradley~A Mullens.
\newblock Sulfur dust bag: a novel technique for ectoparasite control in
  poultry systems.
\newblock {\em Journal of economic entomology}, 109(5):2229--2233, 2016.

\bibitem{murillo2017review}
Amy~C Murillo and Bradley~A Mullens.
\newblock A review of the biology, ecology, and control of the northern fowl
  mite, ornithonyssus sylviarum (acari: Macronyssidae).
\newblock {\em Veterinary parasitology}, 246:30--37, 2017.

\bibitem{FAO}
OECD/FAO.
\newblock Oecd-fao agricultural.

\bibitem{okada2010avian}
Hironao Okada, Koutarou Suzuki, Tsukamoto Kenji, and Toshihiro Itoh.
\newblock Avian influenza surveillance system in poultry farms using wireless
  sensor network.
\newblock In {\em 2010 Symposium on Design Test Integration and Packaging of
  MEMS/MOEMS (DTIP)}, pages 253--258. IEEE, 2010.

\bibitem{palumbo2016human}
Filippo Palumbo, Claudio Gallicchio, Rita Pucci, and Alessio Micheli.
\newblock Human activity recognition using multisensor data fusion based on
  reservoir computing.
\newblock {\em Journal of Ambient Intelligence and Smart Environments},
  8(2):87--107, 2016.

\bibitem{parmezan2015study}
Antonio Rafael~Sabino Parmezan and Gustavo~EAPA Batista.
\newblock A study of the use of complexity measures in the similarity search
  process adopted by knn algorithm for time series prediction.
\newblock In {\em 2015 IEEE 14th International Conference on Machine Learning
  and Applications (ICMLA)}, pages 45--51. IEEE, 2015.

\bibitem{rakthanmanon2012searching}
Thanawin Rakthanmanon, Bilson Campana, Abdullah Mueen, Gustavo Batista, Brandon
  Westover, Qiang Zhu, Jesin Zakaria, and Eamonn Keogh.
\newblock Searching and mining trillions of time series subsequences under
  dynamic time warping.
\newblock In {\em Proceedings of the 18th ACM SIGKDD international conference
  on Knowledge discovery and data mining}, pages 262--270, 2012.

\bibitem{reiss2011exploring}
Attila Reiss, Markus Weber, and Didier Stricker.
\newblock Exploring and extending the boundaries of physical activity
  recognition.
\newblock In {\em 2011 IEEE International Conference on Systems, Man, and
  Cybernetics}, pages 46--50. IEEE, 2011.

\bibitem{siegford2016assessing}
Janice~M Siegford, John Berezowski, Subir~K Biswas, Courtney~L Daigle, Sabine~G
  Gebhardt-Henrich, Carlos~E Hernandez, Stefan Thurner, and Michael~J Toscano.
\newblock Assessing activity and location of individual laying hens in large
  groups using modern technology.
\newblock {\em Animals}, 6(2):10, 2016.

\bibitem{ciwf}
The~Poultry Site.
\newblock About chickens.

\bibitem{ThePoultrySite}
The~Poultry Site.
\newblock Global poultry trends - developing countries main drivers in chicken
  consumption.

\bibitem{smythe2015behavioral}
Brandon~G Smythe, Jimmy~B Pitzer, Mark~E Wise, Andres~F Cibils, Dawn
  Vanleeuwen, and Ronnie~L Byford.
\newblock Behavioral responses of cattle to naturally occurring seasonal
  populations of horn flies (diptera: Muscidae) under rangeland conditions.
\newblock {\em Journal of economic entomology}, 108(6):2831--2836, 2015.

\bibitem{tixier2011chicken}
Michele Tixier-Boichard, Bertrand Bedhom, and Xavier Rognon.
\newblock Chicken domestication: from archeology to genomics.
\newblock {\em Comptes rendus biologies}, 334(3):197--204, 2011.

\bibitem{walton2018evaluation}
Emily Walton, Christy Casey, Jurgen Mitsch, Jorge~A Vazquez-Diosdado, Juan Yan,
  Tania Dottorini, Keith~A Ellis, Anthony Winterlich, and Jasmeet Kaler.
\newblock Evaluation of sampling frequency, window size and sensor position for
  classification of sheep behaviour.
\newblock {\em Royal Society open science}, 5(2):171442, 2018.

\bibitem{wang2021balancing}
Yidi Wang, Mohsen Karimi, Yecheng Xiang, and Hyoseung Kim.
\newblock Balancing energy efficiency and real-time performance in gpu
  scheduling.
\newblock In {\em 2021 IEEE Real-Time Systems Symposium (RTSS)}, pages
  110--122. IEEE, 2021.

\bibitem{yeh2017matrix}
Chin-Chia~Michael Yeh, Nickolas Kavantzas, and Eamonn Keogh.
\newblock Matrix profile vi: Meaningful multidimensional motif discovery.
\newblock In {\em 2017 IEEE international conference on data mining (ICDM)},
  pages 565--574. IEEE, 2017.

\bibitem{yeh2016matrix}
Chin-Chia~Michael Yeh, Yan Zhu, Liudmila Ulanova, Nurjahan Begum, Yifei Ding,
  Hoang~Anh Dau, Diego~Furtado Silva, Abdullah Mueen, and Eamonn Keogh.
\newblock Matrix profile i: all pairs similarity joins for time series: a
  unifying view that includes motifs, discords and shapelets.
\newblock In {\em 2016 IEEE 16th international conference on data mining
  (ICDM)}, pages 1317--1322. Ieee, 2016.

\end{thebibliography}

\appendix

\end{document}